%% file: paper.tex
\documentclass[
reprint,
amsmath,
amssymb,
aps,
prapplied]{revtex4-2}

\usepackage{graphicx}
\usepackage{xcolor}
\usepackage{caption}
\usepackage{subcaption}
\usepackage{amsmath}
\usepackage{algorithm}
\usepackage{algpseudocode}
\usepackage{enumerate}
\usepackage{tabularx}

\allowdisplaybreaks

\newcommand{\ps}{p_{\mathrm{s}}(i)}
\newcommand{\pses}{\hat{p}_{\mathrm{s}}(i)}
\newcommand{\plo}{p_{\mathrm{s}}^{-}(i)}
\newcommand{\pu}{p_{\mathrm{s}}^{+}(i)}
\newcommand{\per}{\epsilon_p(i)}

\newcommand{\nlbound}{L(\pses)}
\newcommand{\pw}{W(i)}
\newcommand{\pwr}{W_r(i)}
\newcommand{\za}{z_{\alpha}}
\newcommand{\zal}{z_{\alpha}}
\newcommand{\rc}{R_{c}(i)}
\newcommand{\rcl}{R_{c}^{-}(i)}
\newcommand{\rcu}{R_{c}^{+}(i)}
\newcommand{\rces}{\hat{R_{c}}(i)}
\newcommand{\rcer}{e^{\mathrm{max}}(\hat{R_{c}}(i))}
\newcommand{\eff}{e_{\mathrm{itr}}}
\newcommand{\CETS}{\text{CETS}_{c}(i)}

\newcommand{\CETSopt}{\text{CETS}_{c}^{\mathrm{opt}}}
\newcommand{\CETSer}{e(\hat{\text{CETS}}_{c}(i))}
\newcommand{\CIa}{\text{CI}_\alpha}
\newcommand{\hhat}[1]{\ \hat{\rule{0ex}{1.2ex}\mkern-3mu #1}}
\newcommand{\errt}{e_T}
\newcommand{\ninit}{n_{\mathrm{init}}}
\newcommand{\ntar}{n_{\mathrm{tar}}}
\newcommand{\opt}{\mathtt{A}}

\newcommand{\highlight}[1]{\textcolor{black}{#1}}

\algnewcommand\algorithmicinput{\textbf{Input:}}
\algnewcommand\algorithmicoutput{\textbf{Output:}}
\algnewcommand\Input{\item[\algorithmicinput]}%
\algnewcommand\Output{\item[\algorithmicoutput]}%

\begin{document}

\title{A Statistical Analysis for Per-Instance Evaluation of Stochastic Optimizers: \newline Avoiding Unreliable Conclusions}

\author{Moslem~Noori}
\affiliation{1QB Information Technologies (1QBit), Vancouver, British Columbia, Canada}

\author{Elisabetta~Valiante}
\affiliation{1QB Information Technologies (1QBit), Vancouver, British Columbia, Canada}

\author{Thomas~Van~Vaerenbergh}
\affiliation{Hewlett Packard Labs, Hewlett Packard Enterprise, Milpitas, California, USA}

\author{Masoud~Mohseni}
\affiliation{Hewlett Packard Labs, Hewlett Packard Enterprise, Milpitas, California, USA}

\author{Ignacio~Rozada}
\thanks{{\vskip-10pt}{\hskip-9pt}Corresponding author: \href{mailto:ignacio.rozada@1qbit.com}{ignacio.rozada@1qbit.com}\\}
\affiliation{1QB Information Technologies (1QBit), Vancouver, British Columbia, Canada}

\date{\today}
\begin{abstract}
    A key trait of stochastic optimizers is that multiple runs of the same optimizer in attempting to solve the same problem can produce different results. As a result, their performance is evaluated over several repeats, or runs, on the problem. However, the accuracy of the estimated performance metrics depends on the number of runs and should be studied using statistical tools. We present a statistical analysis of the common metrics, and develop guidelines for experiment design to measure the optimizer's performance using these metrics to a high level of confidence and accuracy. To this end, we first discuss the confidence interval of the metrics and how they are related to the number of runs of an experiment. We then derive a lower bound on the number of repeats in order to guarantee achieving a given accuracy in the metrics. Using this bound, we propose an  algorithm to adaptively adjust the number of repeats needed to ensure the accuracy of the evaluated metric. Our simulation results demonstrate the utility of our analysis and how it allows us to conduct reliable benchmarking as well as hyperparameter tuning and prevent us from drawing premature conclusions regarding the performance of stochastic optimizers.
\end{abstract}

\maketitle

\section{Introduction} \label{sec: intro}
\input{intro}

\section{Preliminaries}\label{sec: preliminaries}
\input{preliminaries}

\section{Confidence Interval of the Metrics}\label{sec: confidence}
\input{metrics_confidence}

\section{Choosing the Number of Repeats}\label{sec: num repeats}
\input{sample_size}

\section{Numerical Results} \label{sec: numerical}
\input{numerical_examples}

\section{Conclusions} \label{sec: conclusion}
In our study, we performed a statistical analysis to investigate the effect of the number of repeats on the reliability of the estimated metrics for stochastic optimizers. Using this statistical analysis, we proposed an algorithm to adaptively adjust the number of repeats in an experiment to achieve reliable performance metrics estimates. Numerical results show the utility of our analysis  and the effectiveness of the proposed algorithm in keeping the estimate error withing a given threshold.

The analysis and algorithm that we discuss in this work provide practical guidelines for reliable benchmarking and hyperparameter optimization for stochastic optimizers. The guidelines presented in this paper can help the research community to avoid drawing erroneous conclusions about comparisons of different optimizers, or wasting computational resources to perform unreliable hyperparameter optimization.

\section*{Acknowledgements}
We thank our editor, Marko Bucyk, for his careful review and editing of the manuscript. We thank Helmut G. Katzgraber for helpful feedback. This material is based upon work supported by the Defense Advanced Research Projects Agency (DARPA) through Air Force Research Laboratory Agreement No. FA8650-23-3-7313. The views, opinions, and/or findings expressed are those of the author(s) and should not be interpreted as representing the official views or policies of the Department of Defense or the U.S. Government.

\newpage
\bibliography{metric}

\end{document}

%% file: intro.tex
Exact deterministic optimization methods have the advantage of being able to find the optimal solution for an optimization problem, if such a solution exists. \highlight{Despite this advantage, they are inefficient at solving intractable optimization problems because of their high computational cost~\cite{bianchi2009survey}}. As an alternative to deterministic methods, stochastic optimization methods have been developed that are often more computationally efficient than their deterministic counterparts, at the price of there being no guarantee that a given problem will be solved to optimality~\cite{collet2008stochastic}. Such stochastic methods are quite diverse, ranging from heuristic algorithms on CPUs, GPUs \cite{goto2021high}, and FPGAs~\cite{tsukamoto2017accelerator, goto2019combinatorial, nikhar2024all} to specialized hardware such as quantum~\cite{morita2008mathematical} and custom-designed classical \cite{hizzani2024memristor, moy20221, bhattacharya2024computing, zhang2022qubrim, cilasun20243sat, si2024energy, patel2022logically, yamamoto2017coherent} devices.

A fundamental characteristic of stochastic optimizers is that different repeats, or runs, of the same stochastic optimizer on the same problem yield different results. For example, one of the runs may be lucky and quickly find the solution, whereas another can fail to solve the problem, even after spending significant computational resources. Thus, it is important to define metrics that provide insight into the optimizers' performance and computational complexity, \highlight{while considering their stochastic nature}~\cite{beiranvand2017best, neira2025benchmarking}.

There have been several efforts in the literature to define such metrics. For example, a metric called \emph{computational effort} was proposed by Koza~\cite{koza1994genetic} to measure the performance of genetic algorithms. More recent metrics, such as $R_{99}$ \cite{ronnow2014defining}, \emph{time-to-solution} (TTS) \cite{kowalsky20223},  and \mbox{\emph{energy-to-solution}} (ETS) \cite{pedretti2025solving} have been proposed in the quantum, physics-inspired, and in-memory computing optimization communities, which are essentially special cases of Koza's computational effort metric.

All of these metrics depend on an estimate of the probability of a stochastic optimizer successfully solving a problem. Estimating this probability requires solving the problem with several repeats. Solving the problem with a large number of repeats provides a more accurate estimate of the success probability, but this may not be feasible due to the constrained computational resources or limited access to the solver in the case of special hardware optimizers. On the other hand, using a small number of repeats, while computationally less demanding, could result in an inaccurate estimate of the success probability and of the performance metrics as a result. Thus, it is natural to ask ``How many repeats are enough?'' when seeking to accurately estimate the optimizer's metrics. Answering this question is the focus of this paper.

While this question is often overlooked in some fields, such as quantum and physics-inspired optimization, it has been explored in the genetic programming community. Angeline  concluded~\cite{angeline1996investigation} that Koza's computational effort metric is a random variable and should be handled with proper statistical tools. Later, the authors of Ref.~\cite{christensen2002analysis} showed that Koza's metric underestimates the true computational effort, and the magnitude of this underestimation decreases when increasing the run count. The authors also suggested that some earlier results reported in the literature may not be reproducible due to the low number of repeats. In another work~\cite{keijzer2001adaptive}, it was shown that the confidence interval for Koza's metric is highly volatile, even for moderate success probabilities. \highlight {For example, for a solver with a success probability of $0.37$, the width of the confidence interval calculated using a bootstrapping size of 10,000 is of the same magnitude as the estimated computational effort.} The authors of Ref.~\cite{barrero2015study} even suggested avoiding the use of Koza's computational effort metric, and to use alternative metrics instead.

\highlight{In this paper, we conduct a thorough statistical analysis of performance metrics for stochastic optimizers}, and the effect of the number of repeats on the metrics. Our work has been inspired by the increased interest in emerging stochastic hardware devices such as quantum computers~\cite{horowitz2019quantum} and in-memory~\cite{verma2019memory} and neuromorphic~\cite{schuman2022opportunities} computing devices, among others. Due to their limited availability, it is vital to choose the right number of repeats for an accurate estimate of their performance.

To begin, we formally define a general performance metric, called the \emph{computational effort to solution} (CETS), based on the success probability of a stochastic optimizer. The existing metrics, such as Koza's computational effort metric, $R_{99}$, and TTS, are special cases of CETS. We discuss the confidence interval for the estimate of the success probability and how it is related to the number of repeats. We then explain how the uncertainty in the estimation of the success probability translates into the uncertainty in the metrics and derive their confidence interval. We use this confidence interval to find a lower bound on the number of repeats to guarantee achieving a given accuracy of estimate for the metrics. Furthermore, we propose an algorithm to adaptively adjust the number of repeats in an experiment to accurately estimate the metrics. \highlight{We also discuss the effect of the number of repeats when optimizing CETS. A reliable CETS optimization is essential in choosing the right strategy for restarting the stochastic optimizer, given that computational resources are fixed.}

\highlight{We illustrate the utility of our analysis and proposed algorithm for benchmarking and hyperparameter optimization of stochastic optimizers using numerical results. In our experiments, we consider both synthetic and real data obtained from solving instances from various families of Boolean satisfiability (SAT) problems using a variant of the WalkSAT algorithm~\cite{hoos2000local}. The synthetic data helps with abstracting out both the problem and the optimizer, and focusing solely on the statistical results and showing the generality of our analysis. On the other hand, the data from the WalkSAT solver shows the applicability of our analysis to stochastic optimizers when solving SAT instances. Note that, while we present the results for SAT problems and a specific variant of WalkSAT, our analysis applies to a wide range of optimization problems and stochastic optimizers where the outcome takes the form of a Bernoulli trial, that is, whether the problem has been solved or not. This includes, for example, the quadratic assignment, the travelling salesman, vehicle routing, and quadratic unconstrained binary optimization problems; and classical, quantum, and hybrid stochastic optimizers.}

The paper is organized as follows. Section~\ref{sec: preliminaries} provides an introduction to performance metrics and the statistical tools we use in this work. The derivation of the confidence interval for the success probability and the CETS metric is presented in Section~\ref{sec: confidence}. The lower bound on the number of repeats and our proposed adaptive algorithm are discussed in Section~\ref{sec: num repeats}. Numerical results showing the utility of our statistical analysis and the proposed algorithm are presented in Section~\ref{sec: numerical}, and Section~\ref{sec: conclusion} concludes the paper.

\highlight{Table~\ref{tab: notation summary} summarizes the main notation and performance metrics used throughout the paper. It is included to help readers quickly reference symbols and definitions without having to consult the text repeatedly. Note that the point estimates of the metrics are denoted by the hatted version of their notation in Table~\ref{tab: notation summary}.}

\begin{table}[h]
    \begin{center}
    \resizebox{\linewidth}{!}{%
    \begin{tabular}{|c|c|}
         \hline
         Notation & Definition \\ \hline
         $n$ & Number of repeats \\ \hline
         $i$ & Number of iterations \\ \hline
         $c$ & Confidence in the metric estimate \\ \hline
         $\za$ & $(1 - \alpha / 2)$ critical value of the  normal distribution \\ \hline
         $\CIa$ & $(1 - \alpha)$ confidence interval \\ \hline
         $\eff$ & Effort per optimizer's iterations \\ \hline
         $\ps$ & Success probability after $i$ iterations \\
         \hline
         $\rc$ & Repeats to solve a problem after $i$ iterations with confidence $c$ \\ \hline
         $\CETS$ & Computational effort to solution with $i$ iterations and confidence $c$ \\ \hline
         $\CETSopt$ & Optimal computational effort to solution \\ \hline
         ITS & Iterations-to-solution \\ \hline
         $\per$ & Error in the estimate of $\ps$ \\ \hline
         $\pw$ & Width of $\CIa$ of $\ps$ \\ \hline
         $\pwr$ & Relative width of $\CIa$ of $\ps$ \\ \hline
         $\rcer$ & Relative error in the estimate of $\rc$ \\ \hline
         $\errt$ & Threshold to limit $\rcer$ \\ \hline

    \end{tabular}
    }
    \end{center}
    \caption{Overview of notation, along with  definitions.}
    \label{tab: notation summary}
\end{table}

%% file: preliminaries.tex
In this section, we provide an overview of the common metrics to evaluate the performance of stochastic optimizers. We then review some existing statistical results for these metrics.

\subsection{Overview of Performance Metrics} \label{subsec: metrics overview}
There are two main performance aspects when evaluating stochastic optimizers: 
\highlight{
\begin{enumerate}[i)]
    \item \emph{Success} refers to an optimizer's success in finding the target value for the objective function of an optimization problem. The target value can be the optimal solution, the best known solution of the problem, or a solution within a certain optimality gap.
    \item \emph{Computational complexity} refers to the computational resources needed by an optimizer to solve an optimization problem. These resources can be measured based on the solving runtime, energy, etc.
\end{enumerate}
}

There exist metrics that focus on measuring the performance of stochastic optimizers in terms of one or the other of these performance aspects. For example, the success probability~\cite{cai2013improving, beeson2015trac, chmiela2021learning} of an optimizer in solving a problem measures its success. On the other hand, metrics such as the number of evaluations or iterations, or the runtime~\cite{cai2013improving, beeson2015trac, chmiela2021learning}, reflect the computational complexity of the optimize.

For a given stochastic optimizer, the two performance aspects are intertwined, so they cannot be improved simultaneously, that is, one can increase the success probability by allocating more computational resources to the optimizer or develop a lightweight version of it by using fewer computational resources at the cost of having  a lower success probability. This is why, when comparing two optimizers, both aspects should be taken into account for a fair comparison~\cite{mcgeoch2024not}.

There have been proposals in the literature to combine the measurement of success and computational complexity into one unifying metric. Examples include the computational effort introduced by Koza to evaluate the performance of genetic programming algorithms~\cite{koza1994genetic}, TTS \cite{ronnow2014defining}, and optimized TTS \cite{albash2018demonstration} to evaluate the performance of quantum and classical heuristic optimizers. Despite different names, these metrics are fundamentally very similar, as explained below. 

To formally define the computational metrics in this work, let us begin by modelling the optimizer's run on an optimization problem, using a given computational budget, with a Bernoulli trial. For the sake of simplicity, we assume the number of iterations to be a measure of the computational resources used by the optimizer, but the discussion that follows can be generalized to using time or the number of operations as a measure of the computational resources. We denote the optimizer's success probability, after running i for $i$ iterations, by $\ps$. This means that, after $i$ iterations, the optimizer solves the problem or achieves the target value for the objective function of the optimization problem with probability $\ps$, and fails to do so with probability $1 - \ps$. Due to the stochastic nature of the optimizer, it is not guaranteed that the problem will be solved by running the optimizer only once. \highlight{Thus, we are interested in finding the number of times we must run the optimizer on the same problem, denoted by $\rc$, to ensure the target value is reached at least once with a confidence of $c$.} Using a binomial distribution to model the repeats of the Bernoulli trials, achieving a confidence of $c$ after $\rc$ repeats occur if
\begin{equation}
    c \leq 1 - [1 - \ps]^{\rc}.
\end{equation}
This means that the smallest possible value for $\rc$ is
\begin{equation} \label{eq: R_c original}
    \rc = \max\left(\frac{\ln{(1 - c)}}{\ln{(1 - \ps)}}, 1\right).
\end{equation}
Note that if $\ps \geq c$, $\rc = 1$, meaning the problem is fairly easy to solve for the optimizer. For the sake of simplicity and practical relevance, we assume $\ps \leq c$, which simplifies $\rc$ to
\begin{equation} \label{eq: R_c}
    \rc = \frac{\ln{(1 - c)}}{\ln{(1 - \ps)}}
\end{equation}
in what follows. In the literature on quantum optimization, it is often that $c = 0.99$ and the $\rc$ metric is referred to as $R_{99}$~\cite{perdomo2019readiness}. 

Now, if each iteration of the solver takes an $\eff$ amount of computational effort, then the  computational effort to solution (CETS) of the solver after $i$ iterations to attain a confidence of $c$ is defined as 
\begin{equation} \label{eq: CEST def}
    \CETS = i  \eff \rc.
\end{equation}
The above metric is general and covers common metrics as a special case. It is equivalent to Koza's computational effort metric if the goal is to find the optimal solution. \highlight{Also, if $\eff$ refers to the time per iteration, then CETS becomes the TTS metric \cite{kowalsky20223}. Similarly, if $\eff$ refers to the energy per iteration, then CETS becomes the ETS metric~\cite{pedretti2025solving}}.

It is worth mentioning that CETS captures both the success and computational complexity of stochastic optimizers. Since  these two aspects have an opposite effect on CETS, it is natural to define the following optimization problem to find the minimum CETS:
\begin{align} \label{eq: opt CEST}
    \CETSopt & = \min_{i} \CETS \\
             & = \eff \min_{i} i  \frac{\ln{(1 - c)}}{\ln{(1 - \ps)}}, \nonumber
\end{align}
where $\CETSopt$ denotes the optimized CETS. \highlight{Note that here, we assume a fixed and deterministic $\eff$.  If $\eff$ expresses time-varying or probabilistic behaviour, while the principals of our analysis hold true, changes are needed. For example, if $\eff$ is non-deterministic, the optimization problem~\eqref{eq: opt CEST} becomes a stochastic optimization problem.}

The solution to problem~\eqref{eq: opt CEST} provides insight as to whether it is better to run the optimizer for many low-success repeats, which we call the \emph{fail-fast} mode, or to run it for a few high-success repeats, which we call the \emph{patient} mode, or somewhere in between. For example, if the optimizer lacks an efficient mechanism for escaping local minima, given a total computational budget, it may be more effective to restart the optimizer rather than increase the number of iterations in a single run.

All of the above is straightforward if the $\ps$ of the optimizer is known. This is, however, not the case, and we must first estimate $\ps$ to be able to calculate $\CETS$. Estimating $\ps$ of a solver is indeed a binomial proportion estimation (BPE) problem. From this point forward, we use $\pses$ to refer to the point estimate of $\ps$. Similarly, the sampled estimator of $\CETSopt$ is defined as 
\begin{equation} \label{eq: CESTopt estimate}
    \hhat{\CETSopt} = \min_{i} \hhat{\CETS}
\end{equation}
where $\hhat{\CETS} = \eff i  \frac{\ln{(1 - c)}}{\ln{(1 - \pses)}}$. 

\subsection{Overview of Binomial Proportion Estimation}
Binomial proportion estimation is one of the most well-known and methodologically important concepts in statistics. It has been extensively studied in the literature, specifically, the interval estimation of the binomial proportion \cite{brown2001interval}. To this end, several methods have been proposed to estimate the confidence interval for the success probability of a Bernoulli trial, denoted by $p$. \highlight{In our study, given $i$, the outcome of running a stochastic optimizer on a problem is a  Bernoulli trial where $p = \ps$}.

To estimate $p$, the trial is repeated $n$ times, where we observe $n_s$ successes and $n_f = n - n_s$ failures. In such a setting, we are interested in finding the $(1 - \alpha)$ confidence interval $\CIa$ for $p$, meaning
\begin{equation}
    P(p \in \CIa) \geq 1 - \alpha.
\end{equation}
In what follows, we denote the $(1 - \alpha/2)$ critical value of the normal distribution, \highlight{that is, the point on the horizontal axis of the standard normal curve that leaves an upper tail area of $\alpha/2$ and a lower cumulative probability of $(1 - \alpha/2)$}, by $\zal$.

Several methods have been proposed for finding the $\CIa$ \cite{brown2001interval}. These confidence intervals are often compared based on their coverage and width. Coverage is defined as $C(p, n) = P(p \in \CIa)$ and measures the reliability of the interval including $p$. \highlight{For a reliable estimate of $\CIa$, we should have $C(p, n) \geq 1 - \alpha$, meaning that the nominal confidence level matches the actual coverage probability. This, however, often does not occur, in which case $C(p, n) < 1 - \alpha$.} The width of a confidence interval reflects how tight or loose it is. It is desirable to have a high coverage and a narrow width to achieve a reliable and precise estimate. However, improving either the coverage or the width often comes at the price of degrading the other, that is, one can increase the coverage by increasing the width of the interval and vice versa.

\highlight{ In this work, we use Agresti--Coull method due to its simplicity and good coverage~\cite{brown2001interval, brown2002confidence, anderson23}. The Agresti--Coull confidence interval is an adjusted version of the Wald interval for BPE, which adds a small correction to the observed counts to improve accuracy and coverage, especially for small sample sizes and at the boundaries, that is, $p=0$ and $p=1$.} Defining $\hhat{n} = n + \zal^2$ and $\hhat{n}_s = n_s + \frac{\zal^2}{2}$, the Agresti--Coull $\CIa$ is
\begin{equation} \label{eq: Agresti-Coull CIa}
    \left(\hhat{p} \pm \frac{\zal}{\sqrt{\hhat{n}}} \sqrt{\hhat{p} (1 - \hhat{p})}\right),
\end{equation}
where $\hhat{p} = \hhat{n}_s / \hhat{n}$.

%% file: metrics_confidence.tex
In this section, we first discuss the effect of the number of repeats on the confidence interval of $\ps$. We then extend the analysis to $\CETS$ and $\CETSopt$. Using statistical analysis, we provide guidelines on choosing the number of repeats needed to arrive at a reliable estimate of $\CETSopt$ in \ref{sec: num repeats}.

\subsection{Number of Repeats and the $\CIa$ of $\ps$}
\begin{figure}
    \centering
    \includegraphics[width=\linewidth]{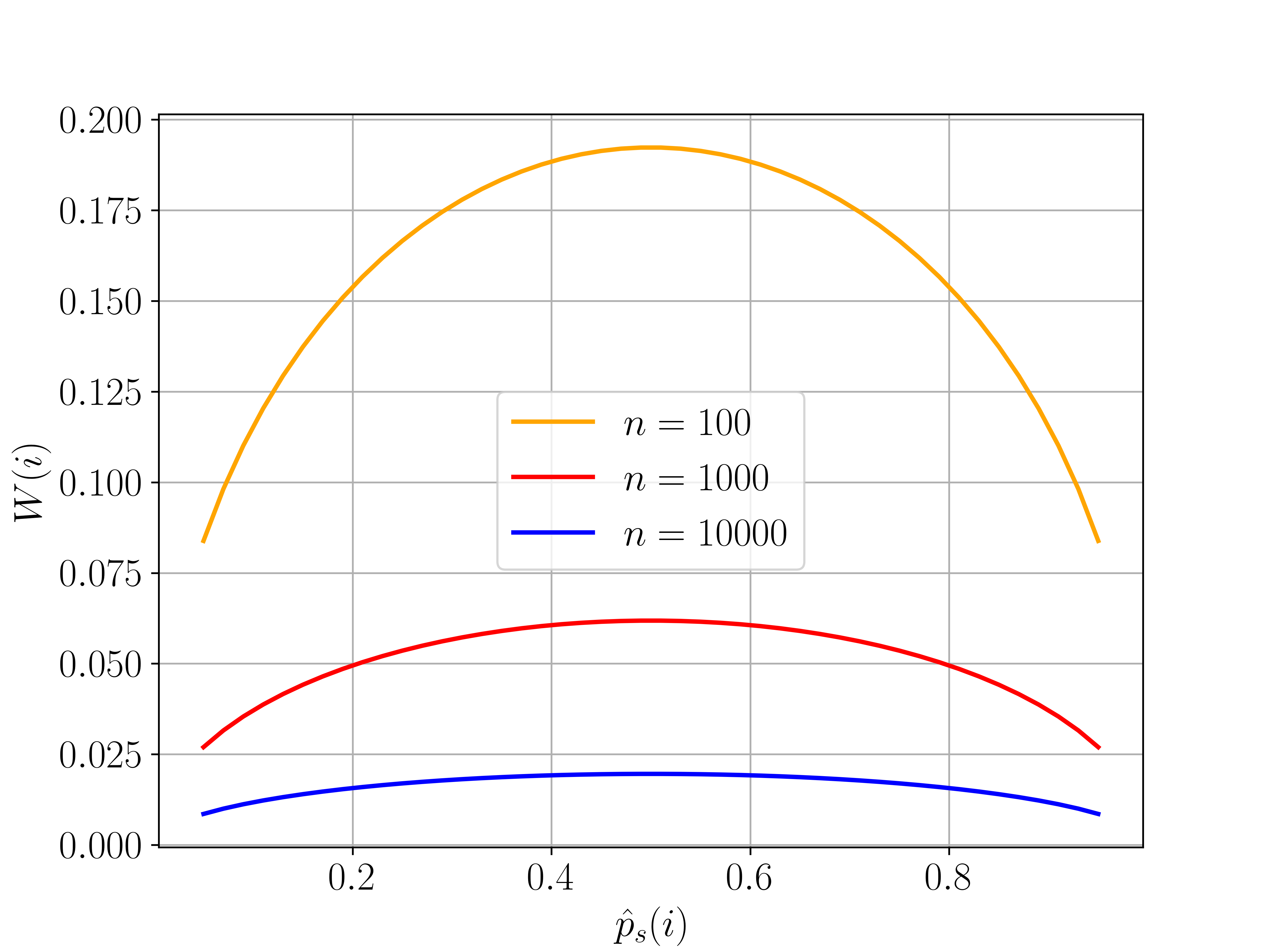}
    \caption{Effect of the number of samples $n$ on the width of the confidence interval of the estimate of $\ps$ for $\alpha = 0.05$.}
    \label{fig:ci_width_abs}
\end{figure}

Choosing the sample size is fundamental in designing experiments to have an accurate BPE~\cite{gonccalves2012sample, krishnamoorthy2007some}. One important metric for choosing the number of Bernoulli trials (or samples) $n$ is the length of the $\CIa$, which reflects the error in the BPE. \highlight{Considering the Agresti--Coull $\CIa$~\eqref{eq: Agresti-Coull CIa}, the width of the $\CIa$ for $\ps$ is}
\begin{equation} \label{eq: CIa width}
    \pw = \frac{2 z_{\alpha}}{\sqrt{\hhat{n}}} \sqrt{\pses [1 - \pses]}.
\end{equation}
From Eq.~\eqref{eq: CIa width}, it is easy to see that for any given $n$, $\pw$ approaches $0$ at $\pses = 0$ and $\pses = 1$, and is maximized at $\pses = 0.5$. Also, intuitively, the estimation of $\ps$ becomes more accurate as $n$ is increased and the estimation error decreases with $1 / \sqrt{n}$. The effect of $n$ on $\pw$ is shown in Fig.~\ref{fig:ci_width_abs}. As shown in this figure, the peak is attained at $\ps = 0.5$ and $\pw$ decreases by an order of magnitude when the sample size is increased from $n = 100$ to $n = $ 10,000.

\begin{figure}
    \centering
    \includegraphics[width=\linewidth]{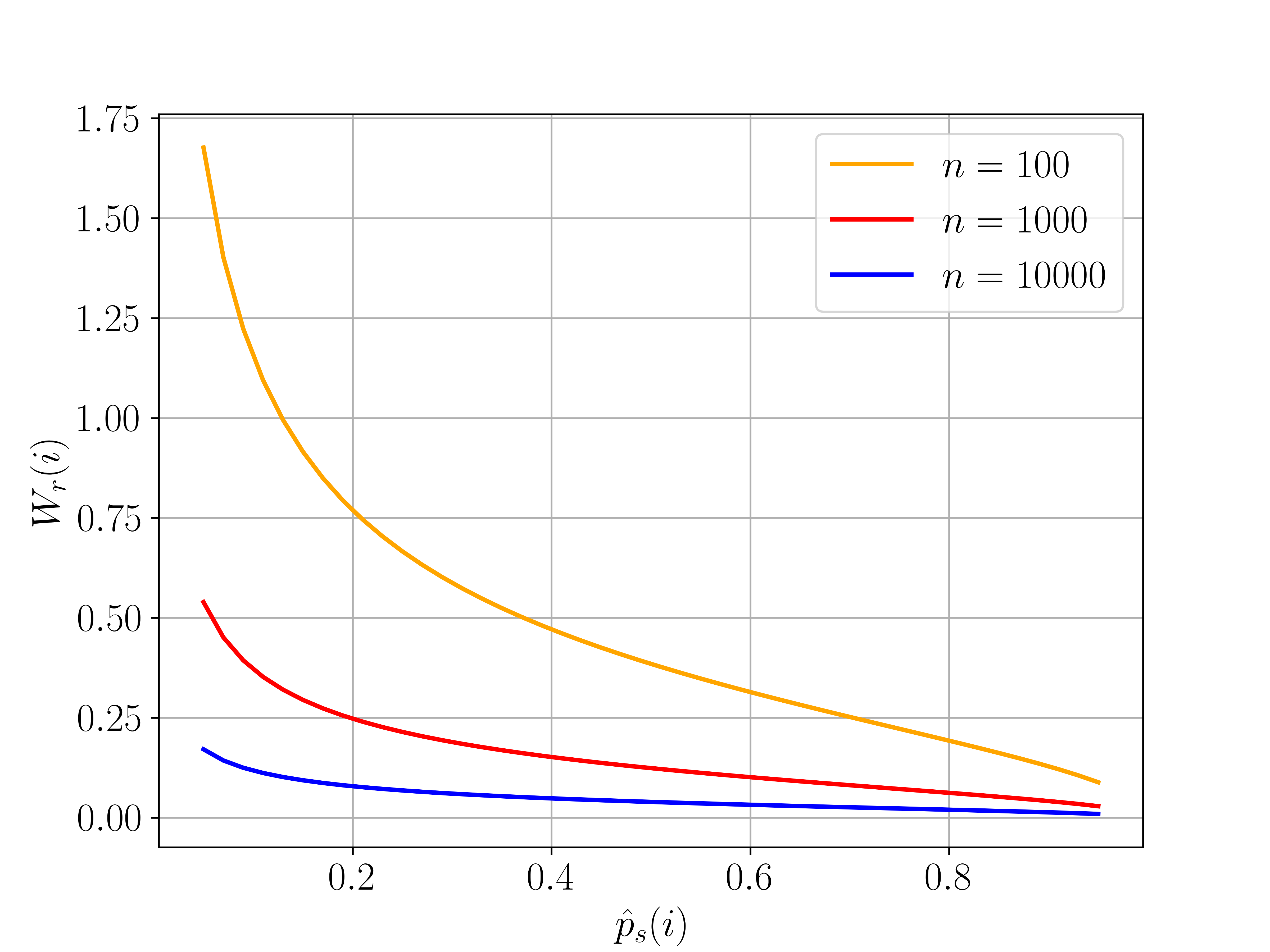}
    \caption{Relative width of the confidence interval compared to the point estimate $\pses$ for three values of $n$ and $\alpha = 0.05$.}
    \label{fig:ci_width_rel}
\end{figure}

The relative width is defined as
\begin{equation} \label{eq: relative CIa width}
    \pwr = \frac{\pw}{\pses} = 2 \zal \sqrt{\frac{ 1 - \pses}{\hhat{n} \pses}},
\end{equation}
and determines the precision of the interval compared to the point estimate $\pses$. A large value of $\pwr$ indicates that the confidence interval is too large compared to the point estimate and the point estimate does not provide a useful insight. It is easy to see from Eq.~\eqref{eq: relative CIa width} that the relative width approaches infinity as $\pses$ approaches $0$ and approaches $0$ as $\pses$ approaches $1$. The relative width decreases with $n$, indicating that a more precise estimate of $\ps$ is achieved by increasing $n$. Figure~\ref{fig:ci_width_rel} depicts the effect of $n$ on $\pwr$.

Now that we have discussed the $\CIa$ for the $\ps$ estimate and its relationship with the number of the optimizer's repeats $n$, we can look at the precision of the $\CIa$ from an experiment design perspective \highlight{for a fixed $i$ and given $\alpha$}. We are interested in finding the number of repeats that we have to run the optimizer on a problem to ensure a maximum estimation error $\per$, that is, $\vert \ps - \pses \vert \leq \per$. This means there should be enough repeats such that $\pw/2 \leq \per$. Substituting $\pw$ from Eq.~\eqref{eq: CIa width} yields
\begin{equation}\label{eq: n bound agresti}
    \hhat{n} = \left \lceil \frac{z_{\alpha}^2 \pses [1 - \pses]}{\per^2} \right \rceil.
\end{equation}
Note that in Eq.~\eqref{eq: n bound agresti}, $\hhat{n}$ depends on the estimated probability $\pses$. However, we do not know $\pses$ prior to running the experiment; thus, we cannot readily use Eq.~\eqref{eq: n bound agresti}. However, $\pses [ 1- \pses]$ is maximized when $\pses = 0.5$ and we can use this as the worst-case scenario. Considering this worst-case scenario  and that $\hhat{n} = n + z_{\alpha}^2$, it is guaranteed that there will be an error margin of at most $\per$ if 
\begin{equation} \label{eq: bound on n to guarantee e}
    n \geq \left \lceil \left(\frac{z_{\alpha}}{2\per} \right)^2  - z_{\alpha}^2 \right \rceil.
\end{equation}
For the $95\%$ confidence interval and assuming  $z_{0.975} \approx 2$, the inequality~\eqref{eq: bound on n to guarantee e} can be simplified to
\begin{equation}
    n \geq \left \lceil \left(\frac{1}{\per} \right)^2  - 4 \right \rceil.
\end{equation}
\subsection{Statistical Analysis of $\CETS$}
In this section, we discuss how the uncertainty in estimating $\ps$ translates into uncertainty in $\CETS$, and, consequently, $\CETSopt$. Since $i$ and $\eff$ do not play any part in the randomness of $\CETS$, we focus on the statistical analysis of $\rc$. The derivation of the results for $\CETS$ from the results for $\rc$ is straightforward. 

First, let us assume that the $\CIa$ of $\ps$ is $(\plo, \pu)$, where $\plo$ and $\pu$ are the lower bound and upper bound of the $\CIa$, respectively. \highlight{Since $\rc$ is a monotonic function of $\ps$,} the $\CIa$ of $\rc$ is $(\rcl, \rcu)$, where
\begin{align} \nonumber
     &\rcl =  \frac{\ln (1  - c)}{\ln(1 - \pu)}, \\
     &\rcu =  \frac{\ln (1  - c)}{\ln(1 - \plo)}.
\end{align}
As presented in the previous section, the $\CIa$ of $\ps$ is $(\pses - \per, \pses + \per)$, where
\begin{equation} \label{eq: p error}
    \per = \frac{z_{\alpha}}{\sqrt{\hhat{n}}} \sqrt{\pses (1 - \pses)}.
\end{equation}
For such a $\CIa$, the point estimate of $\rc$ is \mbox{$\rces = \ln( 1- c) / \ln(1 - \pses)$.} \highlight{For simplicity of presentation and without loss of generality, we assume that $p$ is not too close to the boundaries $p = 0$ and $p = 1$. Otherwise, $\ps - \per < 0$ or $\ps + \per > 1$, and $(\pses - \per, \pses + \per)$ must be clipped accordingly to be confined within $[0 , 1]$, resulting in an asymmetric $\CIa$ for $\ps$.} Now, let us study the behaviour of the $\CIa$ for $\rc$. 

\begin{figure}
    \centering
    \begin{subfigure}{\linewidth}
    \centering
		\includegraphics[width=0.98\linewidth]{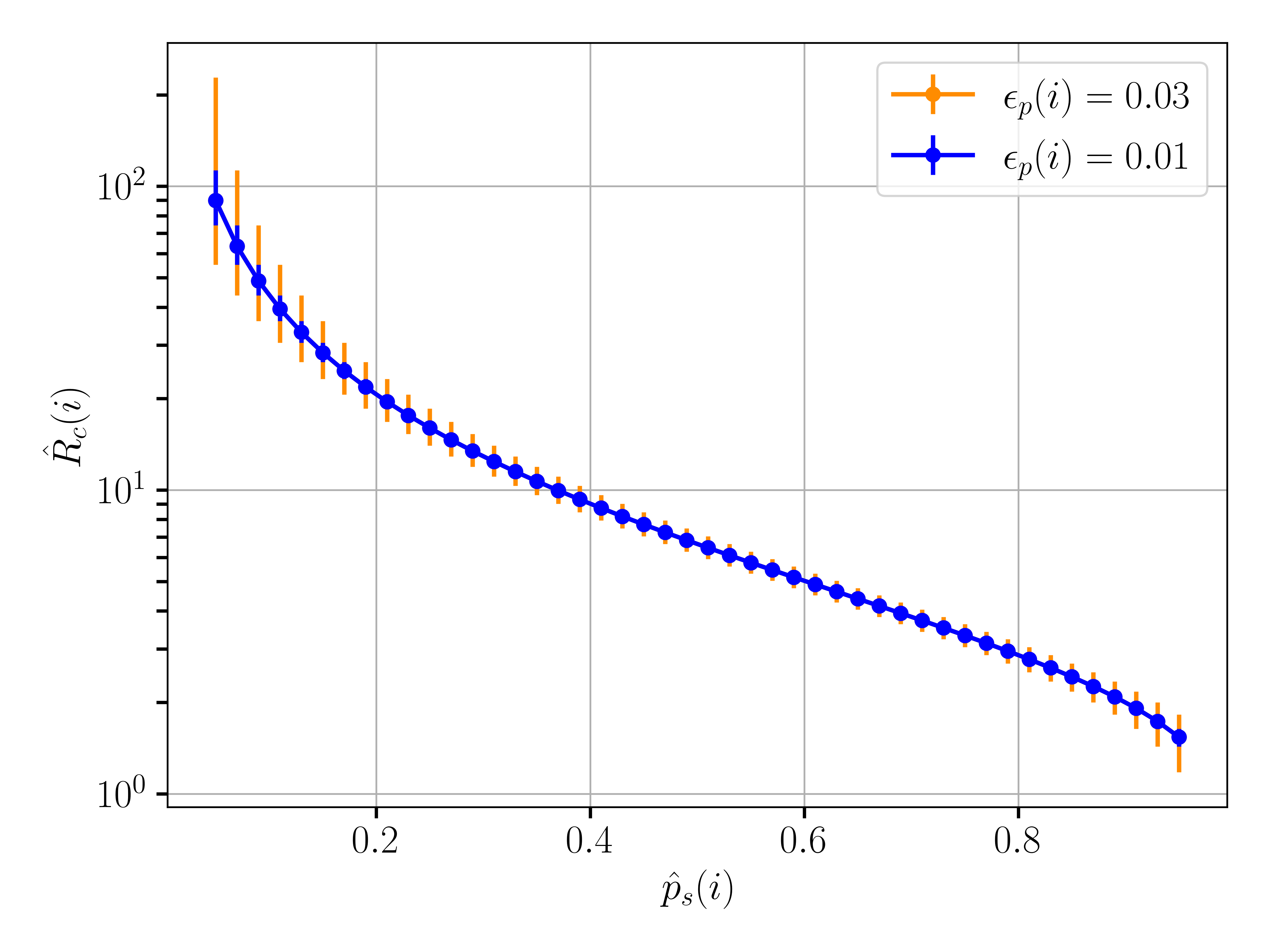}
        \caption{Estimate and confidence interval of $\rc$.}
		\label{fig: r_99 error}        
    \end{subfigure}
    \vspace{1em}

    \centering
    \begin{subfigure}{\linewidth}
    \centering
		\includegraphics[width=\linewidth]{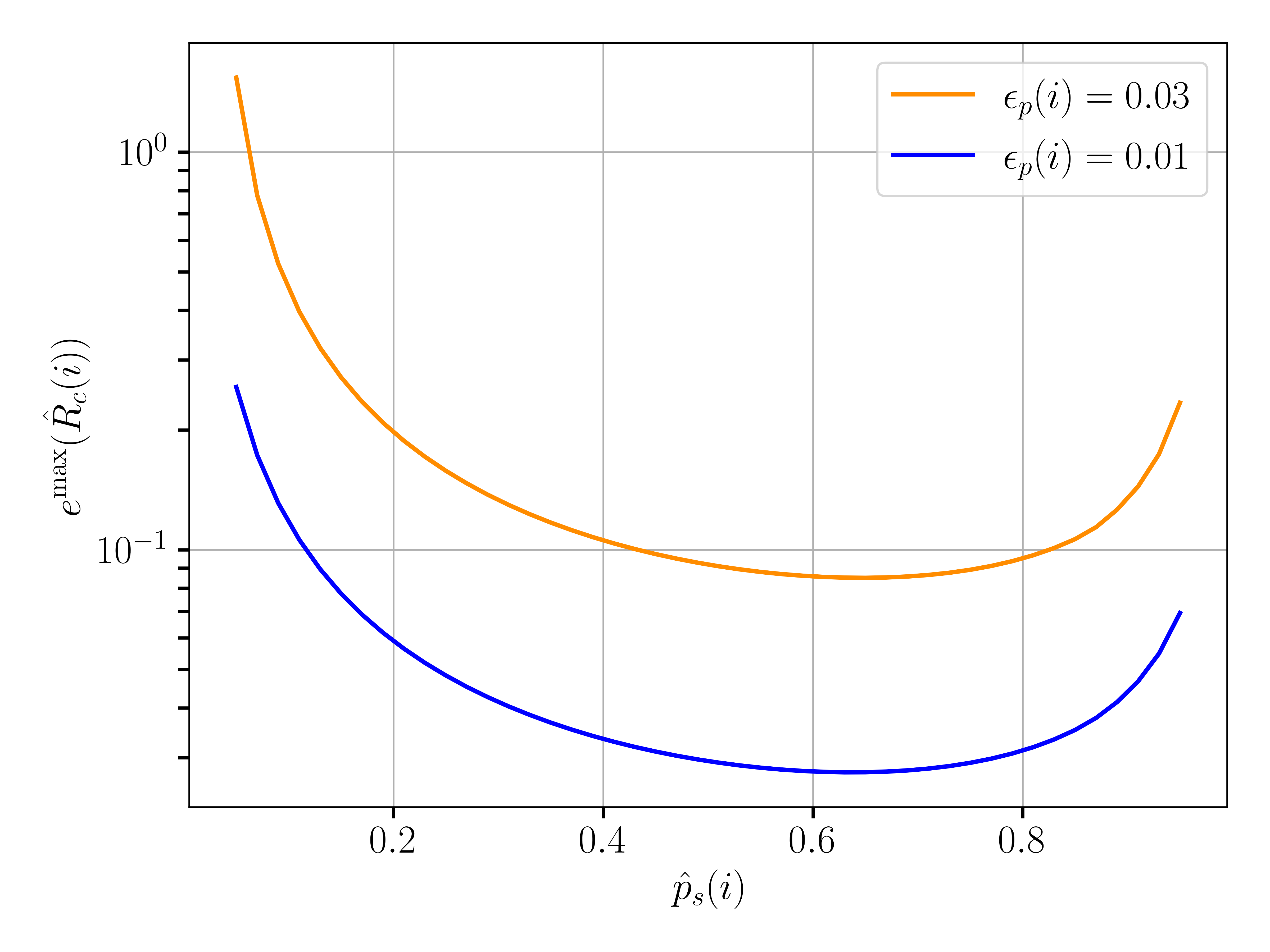}
		\caption{Relative error of the point estimate of $\rc$.}
		\label{fig:max relative error}        
    \end{subfigure}
    \caption{Effect of $\per$ on (a) the  estimate and confidence interval of $\rc$ and (b) the relative estimate error of $\rc$, for $c = 0.99$ and $ \alpha = 0.05$.}
    \label{fig: estimated values}
\end{figure}

While the $\CIa$ is symmetric for $\ps$, it becomes highly asymmetric for $\rc$ due to the nonlinear relationship between $\pses$ and $\rces$. To illustrate this, Fig.~\ref{fig: r_99 error} shows $\rces$ and the $\CIa$ of $\rc$ for $c = 0.99$  when $\pses \in [0.05, 0.96]$, and $\per \in \{ 0.01, 0.03 \}$. As this figure shows, despite having a symmetric $\CIa$ for $\ps$, the $\CIa$ for $\rc$ becomes highly asymmetric, especially as $\pses$ approaches $0$ or $1$ in the case of \mbox{$\per = 0.03$}, due to having chosen a small number of repeats $n$.  

We now look at the precision of the estimate of $\rc$. We define the maximum relative error in the estimate of $\rc$ as 
\begin{equation} \label{eq: relative RC error}
    \rcer = \frac{\max(\rces - \rcl, \rcu - \rces)}{\rces}.
\end{equation}
For the same settings of $\pses$ and $\per$ as in Fig.~\ref{fig: r_99 error}, we plot $\rcer$ in Fig.~\ref{fig:max relative error}. We make an interesting observation from this figure. As expected, $\rcer$ is large for smaller values of the success probability. It then decreases, bottoming out at $\pses \approx 0.6$, and then increases as $\pses$ increases. This differs from the behaviour observed in Fig.~\ref{fig:ci_width_rel}, where $\pwr$ decreases as $\pses$ increases. This illustrates the sensitivity of $\rc$ even for easy problems where $\ps$ is high. 

\begin{figure}
    \centering
    \begin{subfigure}{1.1\linewidth}
    \centering
		\includegraphics[width=\linewidth]{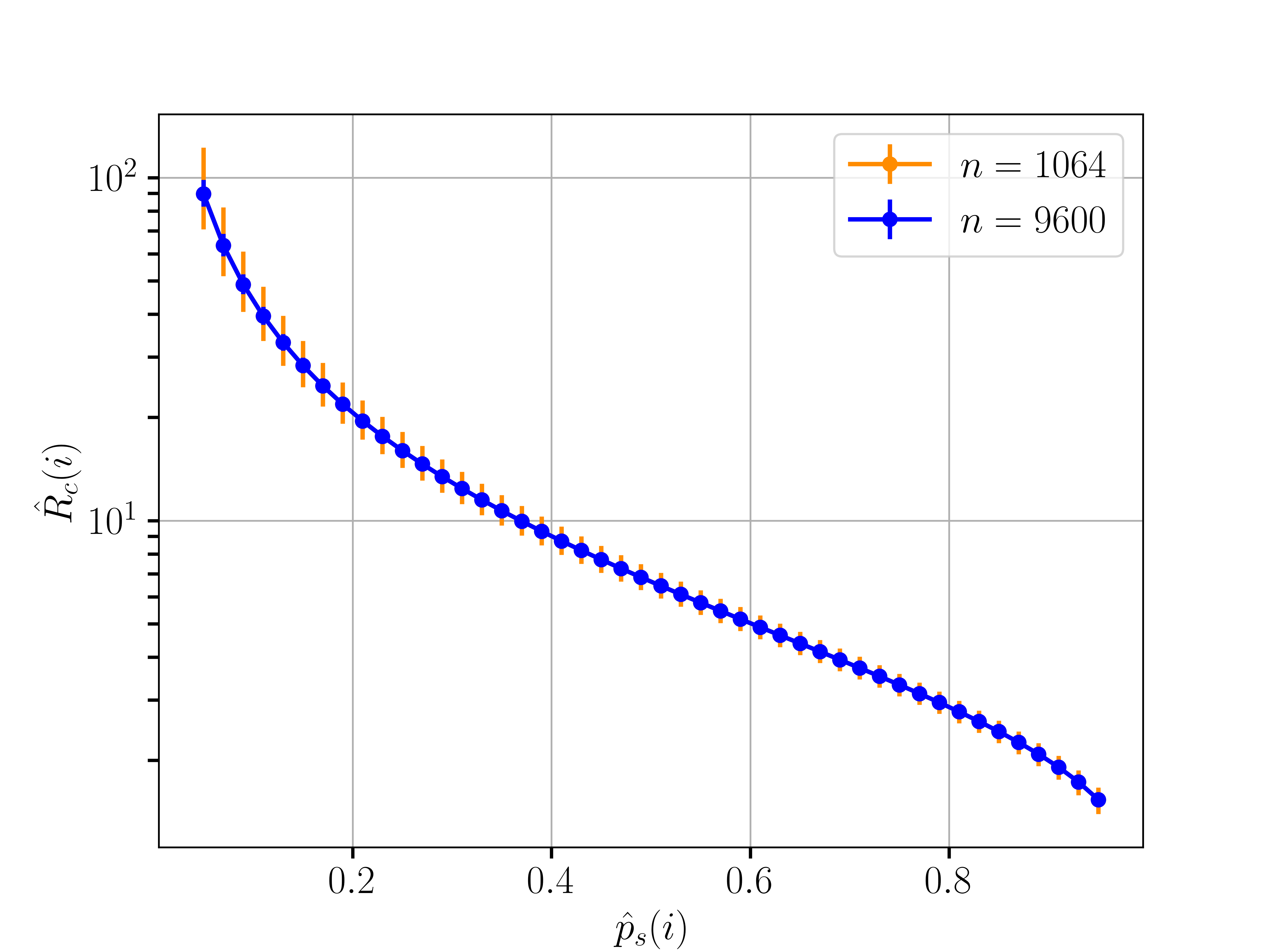}
		\caption{Estimate and confidence interval of $\rc$.}
		\label{fig:Actual rc}
        \vspace{1em}
    \end{subfigure}

    \centering
    \begin{subfigure}{\linewidth}
    \centering
		\includegraphics[width=\linewidth]{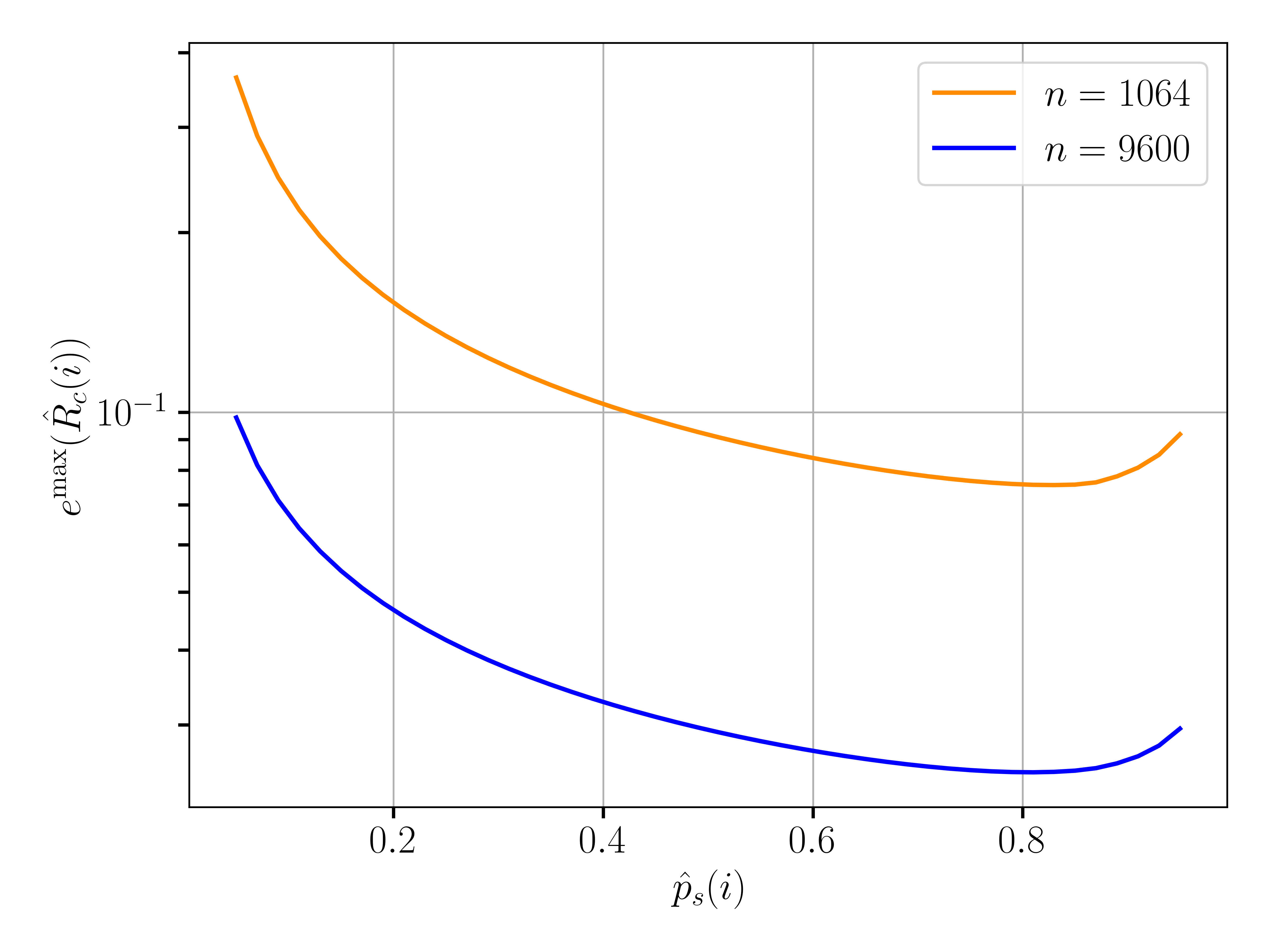}
		\caption{Relative error of the point estimate of $\rc$.}
		\label{fig:actual relative rc}        
    \end{subfigure}
    \vspace{0.5em}
    \caption{Actual (a) estimate and confidence interval of $\rc$ and (b) the relative estimate error of $\rc$, for $c = 0.99$ and $\alpha = 0.05$, when using $n$ derived from the inequality~\eqref{eq: bound on n to guarantee e}.}
    \label{fig:actual values}
\end{figure}

Now that we have shown how the error in estimating $\ps$ propagates into the error in estimating $\rc$, we study the relationship between the number of repeats and $\rces$, as well as its estimate error. We begin by finding the number of repeats needed to \highlight{bound the maximum error to $\per = 0.01$ or $\per = 0.03$.} This is needed when designing an experiment in order to evaluate the performance of the optimizer since we do not know $\pses$ in advance.  Using the inequality~\eqref{eq: bound on n to guarantee e}, we should run the optimizer with $n = 9600$ and $n = 1604$ repeats so that $\per \leq 0.01$ and $\per \leq 0.03$, respectively. Note that the inequality~\eqref{eq: bound on n to guarantee e} is derived for the worst-case scenario, that is, $\pses = 0.5$, and the actual estimation error could be less than $0.01$ and $0.03$ if $\pses \neq 0.5$. Figure~\ref{fig:actual values} shows the actual confidence interval and relative error for $\rc$ across different values of $\pses$ when we run the experiment with $n = 1604$ and $n = 9600$ repeats. While $\per$ reaches its maximum at $\pses = 0.5$, the case is different for $\rcer$. Instead, $\rcer$ increases as $\pses$ approaches $0$ or $1$.  This figure shows that, even with $n = 1064$ repeats and a high success probability (i.e., an easy problem for the optimizer), the estimate can be off by about $10$ percent. The situation is worse for low-success runs (i.e., difficult problems) where the estimate error is about $10$ percent even with $n = 9600$  repeats.

The statistical features of $\CETS$ are similar to those of $\rc$, except with a scaling of $i \eff$. The $\CIa$ of $\CETS$ is $(i \eff \rcl, i \eff \rcu)$ with a point estimate of $i \eff \rces$. The relative error $\CETSer$ in estimating $\CETS$ is equal to that of $\rc$. Furthermore, defining
\begin{equation}
    i^* = \arg\min \CETSer,
\end{equation}
the $\CIa$ and relative estimate error for $\CETSopt$ are those of $\text{CETS}_{c}(i^*)$.

%% file: sample_size.tex
In this section, we present guidelines on designing experiments for evaluating the performance of stochastic optimizers. More specifically, \highlight{fixing the number of iterations $i$}, we discuss how many repeats are enough to ensure a relative error not greater than a given threshold when estimating $\rc$. The discussion is readily applicable when estimating $\CETS$ and $\CETSopt$.

We denote the error threshold by $\errt$, where \mbox{$0 < \errt < 1$}. Note that achieving $\rcer = 0$ requires an infinite number of repeats, which is impractical. We are interested in finding the number of repeats such that
\begin{equation} \label{eq: error condition}
    \rcer \leq \errt.
\end{equation}
Now, considering Eq.~\eqref{eq: relative RC error} and the inequality~\eqref{eq: error condition}, we have
\begin{equation} \label{eq: ineq condition low}
    \frac{\rces - \rcl}{\rces} \leq \errt
\end{equation}
and
\begin{equation} \label{eq: ineq condition high}
    \frac{\rcu - \rces}{\rces} \leq \errt.
\end{equation}
Beginning from the inequality~\eqref{eq: ineq condition low}, we have
\begin{align}\label{eq: ineq condition} \nonumber
    & \frac{\rces - \rcl}{\rces} \leq \errt  \implies \\ \nonumber
    & 1 - \frac{\log(1 - \pses)}{\log(1 - [\pses + \per])} \leq \errt  \implies \\
    & (1 - \errt) \leq \frac{\log(1 - \pses)}{\log(1 - [\pses + \per])}.
\end{align}
Since $\log(1 - [\pses + \per]) < 0$, the final inequality in the sequence of steps~\eqref{eq: ineq condition} results in
\begin{align}\label{eq: ineq condition cont} \nonumber
    & \log(1 - \pses) \leq {(1 - \errt)} \log(1 - [\pses + \per]) \implies \\
    & 1 - \pses \leq (1 - [\pses + \per])^{1 - \errt}.
\end{align}
To find the value of $n$ that satisfies the second inequality in the steps~\eqref{eq: ineq condition cont}, one can first replace $\pses$ by $n_s(i) /(n + \za ^2)$, where $n_s(i)$ is the number of successful repeats after $i$ iterations, and \highlight{replace}  $\per$ from Eq.~\eqref{eq: p error} to express the inequality as a function of $n$. \highlight{Given a value for $\pses$}, a numerical \highlight{optimization} algorithm can be applied to find the smallest value of $n$ that satisfies the inequality~\eqref{eq: ineq condition cont}. As a low-complexity alternative, we use the fact that \mbox{$(1 - x)^a \leq (1 - ax)$} for any $0 \leq a \leq 1$ and $0 \leq x \leq 1$, and simply the inequality to
\begin{equation}
    1 - \pses \leq 1 - (1 - \errt) (\pses + \per),
\end{equation}
which yields
\begin{equation} \label{eq: err bound 2}
    \per \leq \frac{\errt}{1 - \errt} \pses.
\end{equation}
Similarly, beginning from the inequality \eqref{eq: ineq condition high}  and considering that \mbox{$(1 - x)^a \geq (1 - ax)$} for any $1 \leq a $ and $0 \leq x \leq 1$, one can conclude that
\begin{equation} \label{eq: err bound 3}
    \per \leq \frac{\errt}{1 + \errt} \pses.
\end{equation}
Taking the more conservative bound on $\per$, it is guaranteed that $\rcer \leq \errt$ if the inequality~\eqref{eq: err bound 2} is satisfied. \highlight{Substituting $\per$ from the equality~\eqref{eq: p error} into the inequality~\eqref{eq: err bound 3} yields the inequality
\begin{align}\label{eq: lower bound init}
    & \frac{z_{\alpha}}{\sqrt{\hhat{n}}} \sqrt{\pses (1 - \pses)} \leq \frac{\errt}{1 + \errt} \pses.
\end{align}
Defining
\begin{align} \label{eq: bound on n}
    \nlbound =  \left [ \frac{\za (1 + \errt)}{\errt} \right] ^2 \frac{1 - \pses}{\pses} - \za^2,
\end{align}
the inequality~\eqref{eq: lower bound init} results in the following lower bound on the number of repeats to guarantee $\rcer \leq \errt$:
\begin{equation}
    \nlbound \leq n.
\end{equation}
}
Note that, since $(1 + ax)$ is a good approximation for $(1 + x)^a$ for small values of $x$, the above bound is tight for smaller values of $\pses$, while it is loose for larger values of $\pses$.

The main challenge that remains in being able to use \highlight{either the numerical optimization method or} the above bound to find $n$ is that we do not know $\pses$ prior to running the experiment. On the other hand,  since
\begin{equation}
    \lim_{\pses \rightarrow 0} \nlbound = \infty,
\end{equation}
considering the worst-case scenario is also impractical. To address this challenge, we propose Algorithm~\ref{algo: adaptive n} to adaptively adjust the number of repeats. The idea behind this algorithm is to run the experiment with an initial number of repeats $\ninit$ to find an initial estimate for $\pses$. Note that $\ninit$ should be sufficiently large so that the stochastic optimizer solves the input problem at least once. \highlight{For this value of $\pses$, the target number of repeats needed to achieve the desired reliability, denoted by $\ntar$, is calculated  using either $\nlbound$ or the numerical method.} The algorithm then continues by running more repeats and finding a new estimate for $\pses$ and $\ntar$, until $\ntar \leq n$. \highlight{We show in the following section that this algorithm can effectively adjust the number of repeats needed to reach the target error, regardless of the optimizer and the problem.}

\begin{algorithm}[H]
\caption{~~Procedure for Adaptively Adjusting $n$}
\label{algo: adaptive n}
\begin{algorithmic}
\Input \\
\highlight{$\text{P}$, An optimization problem} \\
$\opt$, a stochastic optimizer\\
$\ninit$, the initial number of repeats \\
$\errt$, the relative error threshold \\
$\alpha$, the confidence level

\Output \\
$n$, the number of repeats in the experiment\\
$\pses$, the estimate of the success probability \\
\\
\hspace{-1.25em}
{\bf Procedure:}
\State \highlight{Solve $\text{P}$ using $\opt$ for $\ninit$ repeats}
\State \highlight{Calculate $\pses$ from the result of $\opt$'s $\ninit$ repeats}
\State \highlight{Calculate $\ntar$ using $\nlbound$ or the numerical method}
\State $n \leftarrow \ninit$
\While{ \highlight{$n < \ntar$}}
\State \highlight{Solve $\text{P}$ using additional $\left[ \ntar - n \right]$ repeats}

\State \highlight{$n \leftarrow \ntar$}
\State Update $\pses$ using all collected $n$ repeats
\State \highlight{Update $\ntar$ using $\nlbound$ or the numerical method}
\EndWhile
\State \Return $n, \pses$
\end{algorithmic}
\end{algorithm}

%% file: numerical_examples.tex
In this section, we present numerical results that show the importance of using a sufficient number of repeats and the utility of our experiment design guidelines. For this purpose, we use both synthetic and real data.

\subsection{Relative Error and $\CIa$: Synthetic Data}
Let us begin by considering a scenario where we wish to compare the performance of two stochastic optimizers solving the same problem. The true success probabilities of the first and second optimizers are $p_1$ and $p_2$, respectively, where $p_1 > p_2$. In reality, these probabilities are unknown to us; hence, to compare the solvers, we solve the problem using $n$ repeats, estimate $p_1$ and $p_2$ using these repeats, and then calculate the metric for the optimizers using the estimated values $\hat{p}_1$ and $\hat{p_2}$. In what follows, we use the $R_{99}$ metric (i.e., $c$ = 0.99) and $\alpha = 0.05$ to evaluate the performance of the optimizer, but the observations would be similar for other metrics and different values of~$c$ and $\alpha$.

To synthetically mimic the result of solving the problem by the two optimizers, we generate two random binary vectors, each having the size $n$. The elements of the vectors equal to 1 (or 0) represent the successful (or unsuccessful) repeats. Denoting a Bernoulli distribution with a parameter $p$ by \highlight{$\mathrm{Bern}(p)$}, the first vector is generated using \highlight{$\mathrm{Bern}(p_1)$} and the second vector is generated using \highlight{$\mathrm{Bern}(p_2)$}. Using these vectors, we then find $\hat{p}_1$ and $\hat{p}_2$ and their confidence intervals using the Agresti--Coull method. Ideally, the experiment should find $\hat{p}_1 > \hat{p_2}$, and the estimated $R_{99}$ for the first optimizer, $\hat{R}_{99}^1$, to be smaller than the estimated $R_{99}$ for the second optimizer, $\hat{R}_{99}^2$, that is, $\hat{R}_{99}^1 < \hat{R}_{99}^2$. To investigate how frequently this happens, we run the experiment $1000$ times to collect statistical data for several pairs of $(p_1, p_2)$. The results for $(p_1, p_2) \in \{ (0.25, 0.2), (0.5, 0.45), (0.75, 0.7), (0.99, 0.94) \}$ are presented in Table~\ref{tab: p comparison}.

\begin{table}[]
    \centering
    \begin{tabular}{|c||c|c|c|}
        \hline
        $n$ & 100 & 1000 & 10,000 \\
        \hline \hline
        $(p_1, p_2) = (0.25, 0.2)$ &  &  &  \\
         $\hat{R}_{99}^1 < \hat{R}_{99}^2$ & 0.764 & 0.997 & 1.000 \\
         No $\CIa$ overlap & 0.026 & 0.450 & 1.000 \\
        \hline \hline
        $(p_1, p_2) = (0.5, 0.45)$ &  &  &  \\
         $\hat{R}_{99}^1 < \hat{R}_{99}^2$ & 0.717 & 0.982 & 1.000 \\
         No $\CIa$ overlap & 0.026 & 0.317 & 1.000 \\
        \hline \hline
        $(p_1, p_2) = (0.75, 0.7)$ &  &  &  \\
         $\hat{R}_{99}^1 < \hat{R}_{99}^2$ & 0.790 & 0.991 & 1.000 \\
         No $\CIa$ overlap & 0.028 & 0.379 & 1.000 \\
        \hline \hline
        $(p_1, p_2) = (0.99, 0.94)$ &  &  &  \\
         $\hat{R}_{99}^1 < \hat{R}_{99}^2$ & 0.967 & 1.000 & 1.000 \\
         No $\CIa$ overlap & 0.369 & 1.000 & 1.000 \\
        \hline
    \end{tabular}
        \caption{Comparison of the hypothetical first and second optimizers over different pairs of $(p_1, p_2)$ and different numbers of repeats $n$.}
    \label{tab: p comparison}
\end{table}

The table shows the chance of correctly finding \mbox{$\hat{R}_{99}^1 < \hat{R}_{99}^2$} as well as the chance of having no overlap between the $\CIa$ of $\hat{R}_{99}^1$ and $\hat{R}_{99}^2$ for different pairs of $(p_1, p_2)$. Note that the higher the chance is of there being no overlap between the confidence intervals, the higher the reliability is of the  comparison of the metrics of the two optimizers. Thus, in the ideal case, the chance of each will be $1$. As shown in this table, for \mbox{$n = 100$} and $(p_1, p_2) \in \{ (0.25, 0.2), (0.5, 0.45), (0.75, 0.7) \}$, there is a significant chance of observing $\hat{R}_{99}^1 > \hat{R}_{99}^2$, and consequently arriving at an incorrect conclusion about the performance of the two optimizers, although the high overlap between the confidence intervals of the metrics makes this conclusion unreliable. For $n = 100$ and $(p_1, p_2) \in \{ (0.25, 0.2), (0.5, 0.45), (0.75, 0.7) \}$, even when we observe $\hat{R}_{99}^1 < \hat{R}_{99}^2$, it is almost certain that the confidence intervals of the metrics of the two optimizers overlap, lowering our confidence in the comparison. Even for $(p_1, p_2) = (0.99, 0.94)$, there is  more likely to be an overlap between the $\CIa$ of the $R_{99}$ values than for there to be no overlap. By increasing the number of repeats, it becomes more likely that $\hat{R}_{99}^1 < \hat{R}_{99}^2$. For $n =$ 10,000, the better solver can always be identified with a very high level of confidence.

We now evaluate the utility of Algorithm~\ref{algo: adaptive n} in choosing the number of repeats. For this purpose, we generate synthetic data for different values of a true optimizer's success probability $p$. We again use a Bernoulli distribution \highlight{$\mathrm{Bern}(p)$} to model the outcome of the optimizer's runs. In our experiment, we assume $\ninit  = 100$ and a maximum relative error of $\errt = 0.1$ when running  Algorithm~\ref{algo: adaptive n}. The true $p$ and returned $\pses$ by Algorithm~\ref{algo: adaptive n} are then used to calculate $R_{99}$ and $\hat{R}_{99}$, respectively. The true relative error in estimating $R_{99}$ is calculated using these two numbers \highlight{as follows:
\begin{equation}
    e(\hat{R}_{99}) = \frac{\vert \hat{R}_{99} - R_{99} \vert}{R_{99}}.
\end{equation}
Note that the above equation yields the actual relative estimation error for $\hat{R}_{99}$, given the true value of $R_{99}$, while Eq.~\eqref{eq: relative RC error} defines an upper bound on the estimation error.}

\begin{figure}
    \centering
    \includegraphics[width=\linewidth]{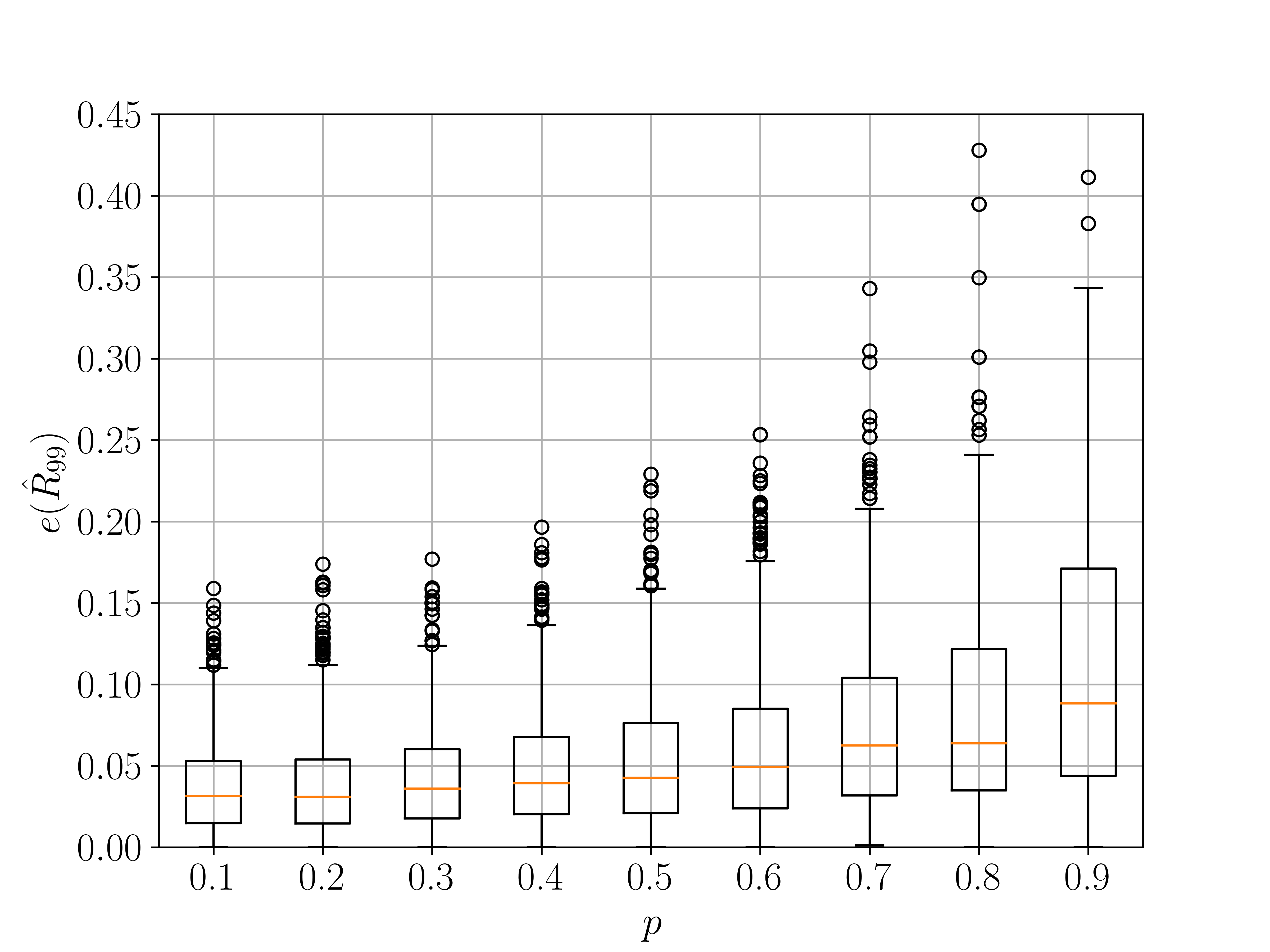}
    \caption{Relative estimate error for different values of the true success probability $p$ resulting from applying Algorithm~\ref{algo: adaptive n} for a target maximum relative error $\errt = 0.1$. \highlight{Circles depict data points that lie farther than 1.5 times the interquartile range (IQR) from the first or third quartile.}}
    \label{fig: relative error}
\end{figure}

To collect statistical data, we run the experiment $1000$ times. The box plot for the relative estimate error of the true $R_{99}$ is presented in Fig.~\ref{fig: relative error}. Figure~\ref{fig: relative error} shows that, for smaller values of $p$, there is a very a high change of the true relative error being below $0.1$. Note that these small values of $p$ belong to the region where \mbox{$(1 + x)^a \approx 1 + ax$} is an accurate approximation, making $\nlbound$ tight when used in Algorithm~\ref{algo: adaptive n}. As we increase $p$, the relative error increases, since $\nlbound$ becomes loose. Despite this behaviour, for $p = 0.9$, the median relative error is still below $0.1$.

\begin{figure}
    \centering
    \includegraphics[width=\linewidth]{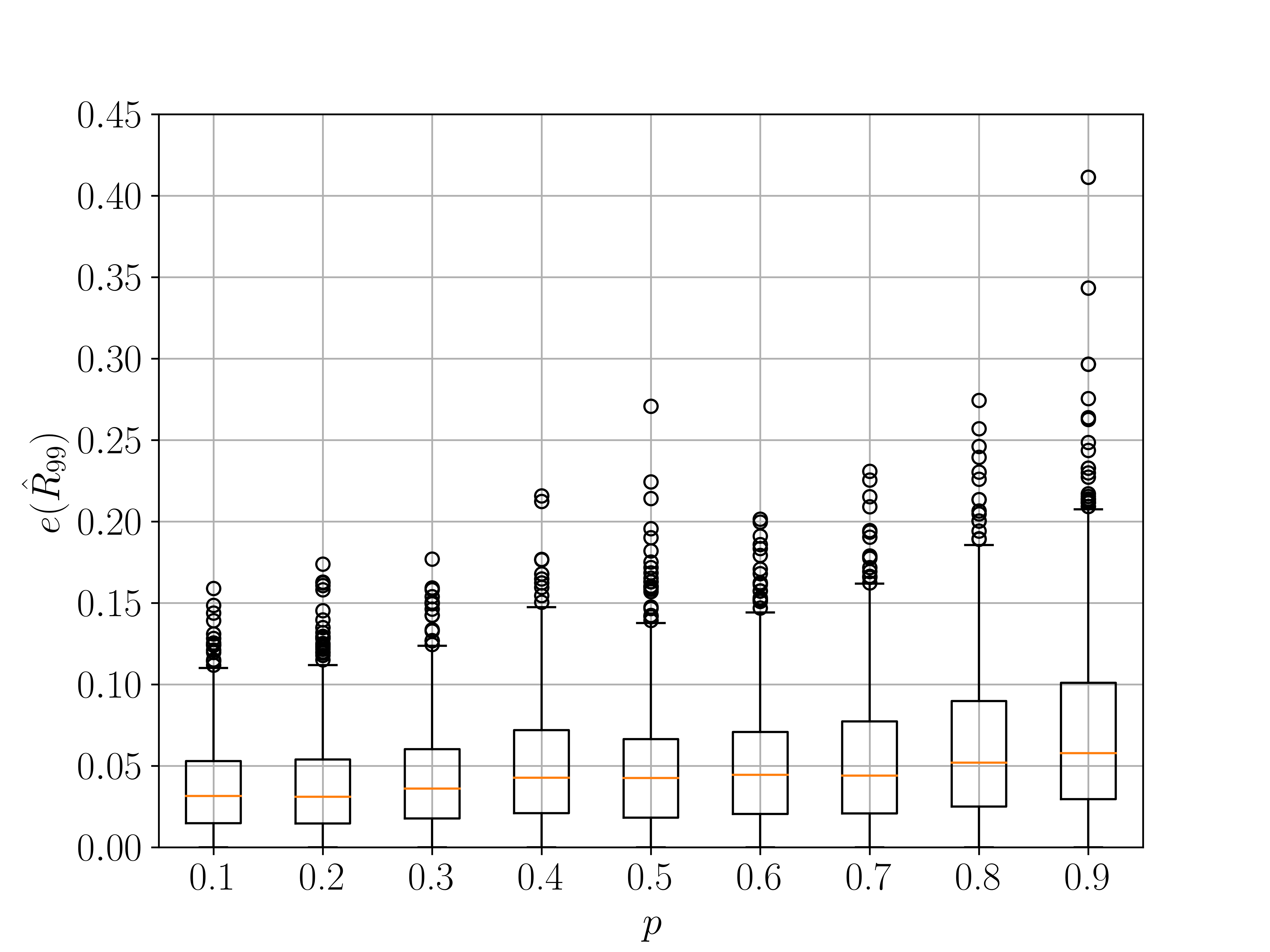}
    \caption{Relative estimate error for different values of the true success probability $p$ resulting from applying Algorithm~\ref{algo: adaptive n} and the modified lower bound for a target maximum relative error $\errt = 0.1$. \highlight{Circles depict data points that lie farther than 1.5 times the interquartile range (IQR) from the first or third quartile.}}
    \label{fig: adaptive n modified}
\end{figure}

\begin{figure}
    \centering
    \includegraphics[width=\linewidth]{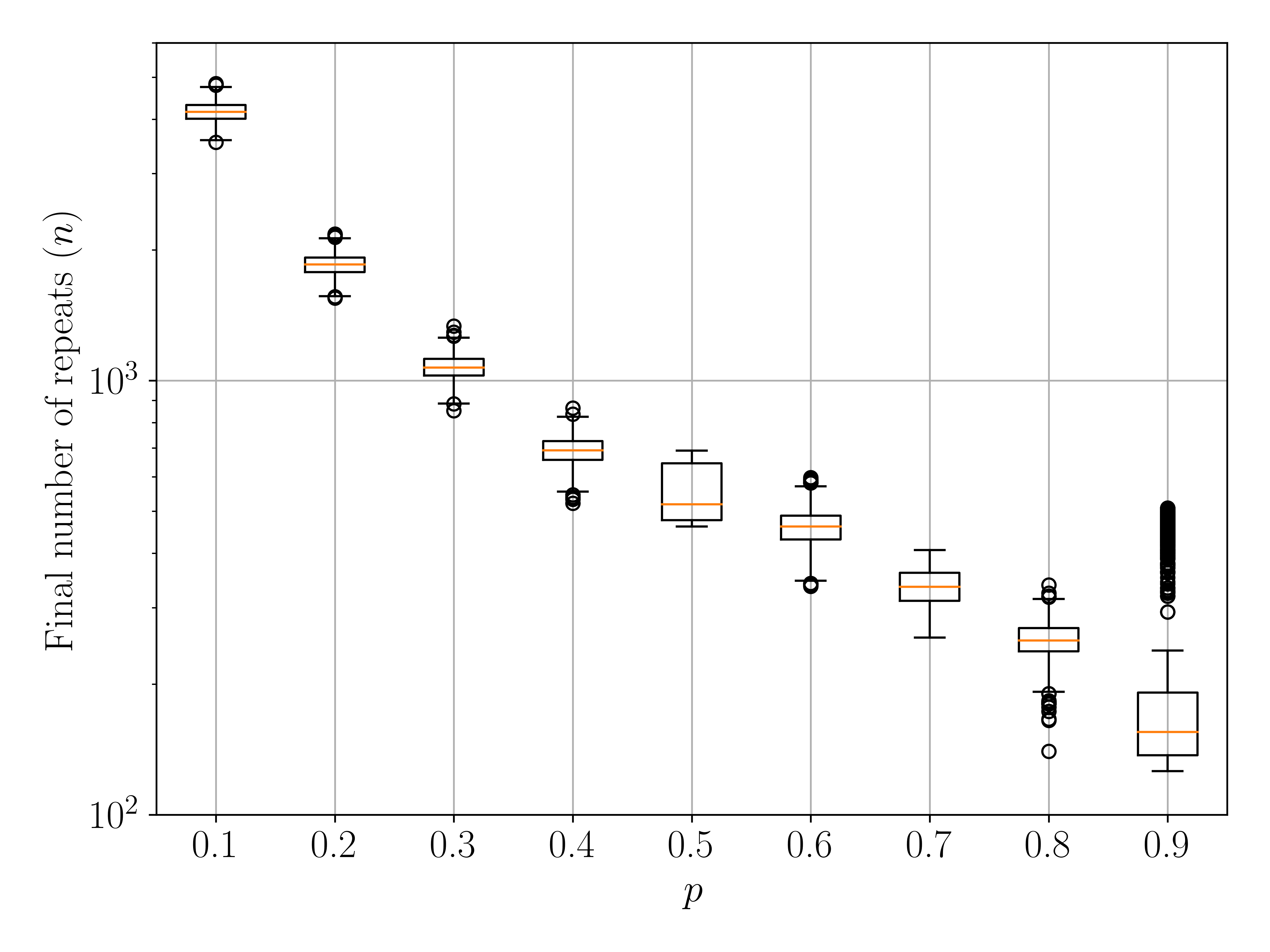}
    \caption{Number of repeats suggested by Algorithm~\ref{algo: adaptive n} needed to limit the relative error to $\errt = 0.1$ using the modified lower bound. \highlight{Circles show data points that lie beyond 1.5 times the interquartile range (IQR) from the first or third quartile.}}
    \label{fig: num repeats algo modified}
\end{figure}

As shown in Fig.~\ref{fig: relative error}, the relative estimate error resulting from adjusting $n$ by using Algorithm~\ref{algo: adaptive n} is higher than the expected threshold for larger values of $p$. This is largely due to the lower bound in Eq.~\eqref{eq: bound on n} not being tight for larger values of $p$. To address this issue, especially when there are stringent conditions on the maximum tolerable relative estimate error, one can either use a root-finding algorithm to find the value of $n$ as explained in Section~\ref{sec: num repeats} or use the following modified lower bound on $n$:
\begin{equation} \label{eq: modified lower bound}
    \nlbound = s(\pses) \left( \! \left[\! \frac{\za (1 + \errt)}{\errt} \! \right] ^2 \! \frac{1 - \pses}{\pses} - \za^2 \! \right),
\end{equation}
where $s(\pses)$ is a scaling function. \highlight{Here, an adjustment term $s(\pses)$ is introduced to make the lower bound more accurate. For smaller values of $\pses$, the bound is tight and $s(\pses) = 1$, while for larger $\pses$ values, $s(\pses)$ should be larger. Empirical or analytical methods can be used to find a function $s(\pses)$ that compensates well for the approximation error. Through numerical experiments, we empirically find that using the following}
\begin{equation}\label{eq: scaler function}
    s(\pses) =
    \begin{cases}
        1 , & \text{if } \pses \leq 0.5,\\
        1.5 , & \text{if } 0.5 < \pses \leq 0.7, \\
        2 , & \text{if } 0.7 < \pses \leq 0.8, \\
        2.5 , & \text{if } 0.8 < \pses \leq 0.9, \\
        10 , & \text{if } 0.9 < \pses \\
    \end{cases}
\end{equation}
can improve the tightness of the lower $\nlbound$. The relative error of $R_{99}$ using the lower bound in Eqs.~\eqref{eq: modified lower bound} and \eqref{eq: scaler function} is presented in Fig.~\ref{fig: adaptive n modified}. As shown in the figure, the results are improved and the error is kept below the threshold in the majority of cases even fo $p = 0.9$. \highlight{The number of repeats resulting from running Algorithm~\ref{algo: adaptive n} is plotted in Fig.~\ref{fig: num repeats algo modified}. As shown in the figure, the number of repeats needed to achieve the $e(\hat{R}_{99})$ values in Fig.~\ref{fig: adaptive n modified} varies from a few hundred for $p = 0.9$ to about $5000$ for $p = 0.1$.}

\begin{figure}
    \centering
    \includegraphics[width=\linewidth]{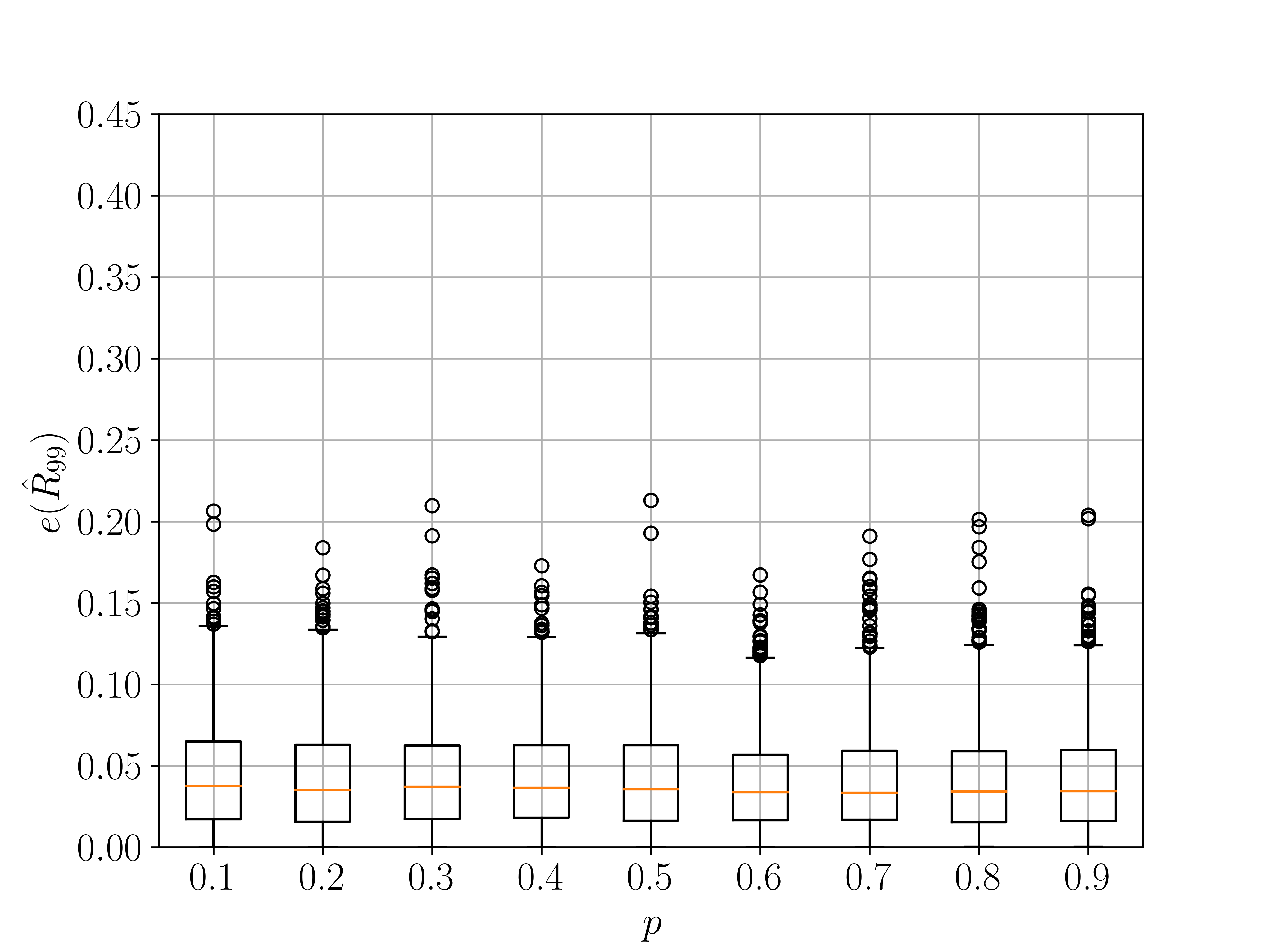}
    \caption{Relative estimate error for different values of the true success probability $p$ resulting from applying Algorithm~\ref{algo: adaptive n} and the exact method for a target maximum relative error $\errt = 0.1$. \highlight{Circles show data points that lie beyond 1.5 times the interquartile range (IQR) from the first or third quartile.}}
    \label{fig: adaptive n exact}
\end{figure}

\begin{figure}
    \centering
    \includegraphics[width=\linewidth]{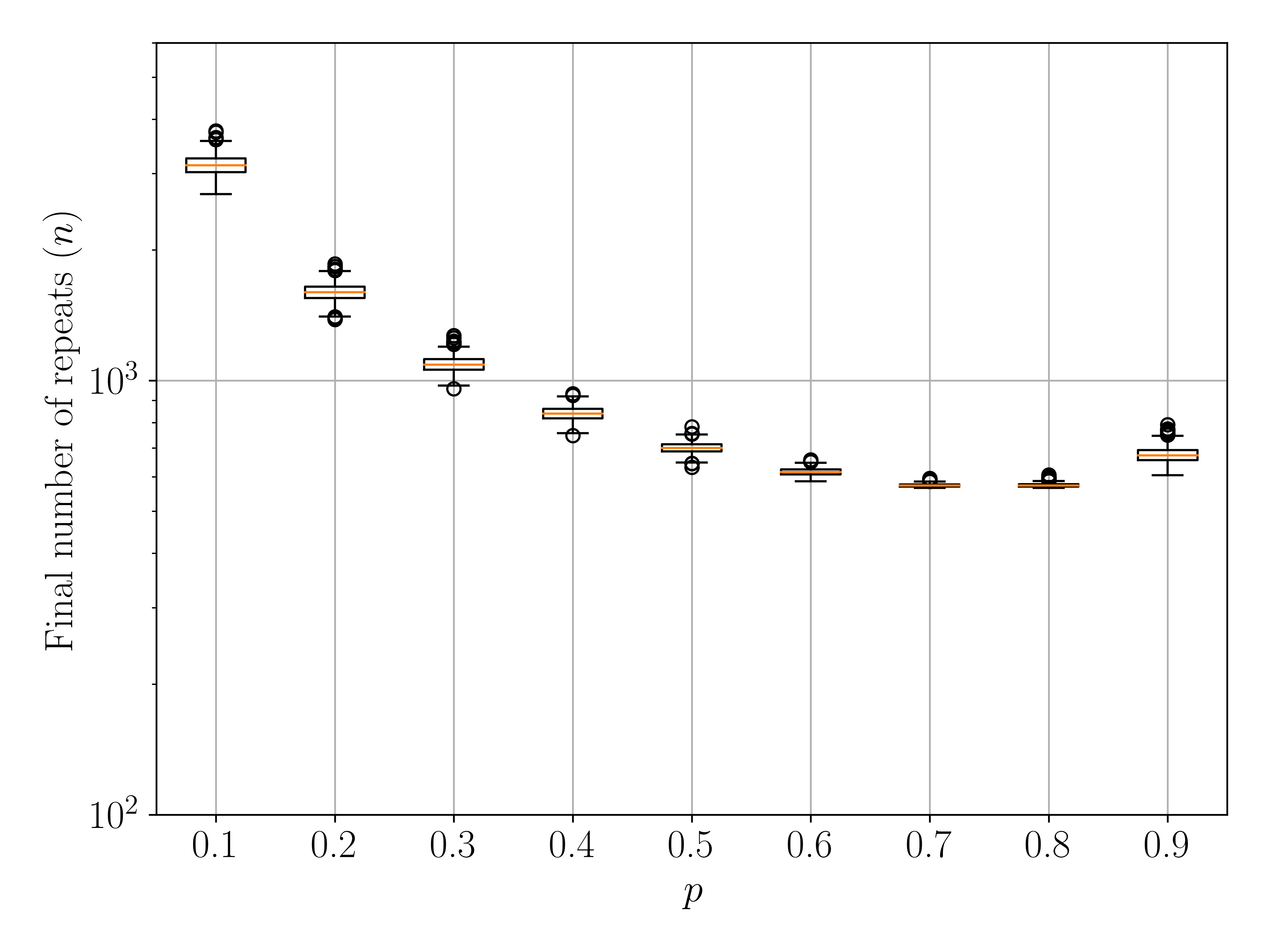}
    \caption{Number of repeats suggested by Algorithm~\ref{algo: adaptive n} needed to limit the relative error to $\errt = 0.1$ using the exact method. \highlight{Circles show data points that lie beyond 1.5 times the interquartile range (IQR) from the first or third quartile.}}
    \label{fig: num repeats algo exact}
\end{figure}

\highlight{To study the effect of approximation used to derive the bound~\eqref{eq: bound on n} on the outcome of Algorithm~\ref{algo: adaptive n}, we run a similar experiment, with the difference being that we use a numerical method to find $n$ by solving the inequality~\eqref{eq: ineq condition cont} instead of using the bound~\eqref{eq: bound on n}. The result of this experiment is presented in Figs.~\ref{fig: adaptive n exact} and \ref{fig: num repeats algo exact}. We see an improvement in the  results for $e(\hat{R}_{99})$, where the relative error is kept below $\errt$ for most of the time, even when $p = 0.9$. There is also a difference in the final number of repeats when Algorithm~\ref{algo: adaptive n} uses the lower bound versus the numerical method. Figure~\ref{fig: num repeats algo exact} shows that, using the numerical method, Algorithm~\ref{algo: adaptive n} requires fewer repeats to converge in the case of smaller values of $p$ and more repeats for larger values of $p$, compared to using the lower bound. For mid-range values of $p$, the number of repeats is similar between the two.}

\highlight{Results such as those shown in Figs.~\ref{fig: adaptive n modified} and \ref{fig: num repeats algo exact} can help us gain insight into the reliability of results reported in prior studies. For example, suppose an earlier work had reported the $R_{99}$ metric for a problem where the optimizer's success probability is about $0.3$. If the number of repeats in that work were reported to be about 1000, the results in these figures indicate that the expected range of $e(\hat{R}_{99})$, with $95\%$ confidence for $\alpha = 0.05$, is about 0.1. If the work reports a smaller number of repeats, for example, $n = 100$, the estimation error is larger and can be found using our analysis in Section~\ref{sec: num repeats}.}

\subsection{Relative Error and $\CIa$: Real Data}
In this section, we show the utility of our analysis applied to results obtained from solving optimization problems.
To do so, we select three families of SAT \mbox{problems \cite{biere2009handbook}} from the \texttt{SATLIB} \cite{SATLIB} library and solve 10 problems from each family. The selected problem families are \texttt{CBS-k3-n100-m435-b10}, \texttt{uf100-430}, and \texttt{uf250-1065}. The three families of problems are selected from various difficulty levels to show the importance of our analysis in different scenarios. We use a variation of the WalkSAT algorithm available at Ref.~\cite{WalkSAT} to solve each problem. WalkSAT is a widely used stochastic local search solver for SAT problems that is often used as a baseline for benchmarking other SAT solvers.

First, we solve each problem with the default WalkSAT walk probability $w=0.5$ using three values for the number of repeats \mbox{$n \in \{ 100, 1000,$ 10,000$\}$.} Since the problem families have different difficulty levels, we choose a different number of WalkSAT iterations for each family, as shown in Table~\ref{tab:num iterations}. For simplicity, we use $p$ to refer to the success probability of WalkSAT in what follows rather than using $\ps$.

\begin{table}[]
    \centering
    \begin{tabular}{|c|c|}
         \hline
         Problem family & Num. iterations \\ \hline
         \texttt{CBS-k3-n100-m435-b10} & 500 \\ \hline
         \texttt{uf100-430} & 1000 \\ \hline
         \texttt{uf250-1065} & 5000 \\ \hline
    \end{tabular}
    \caption{Number of WalkSAT iterations for each problem family.}
    \label{tab:num iterations}
\end{table}

\begin{figure}
    \centering
    \begin{subfigure}{\linewidth}
        \includegraphics[width=\linewidth]{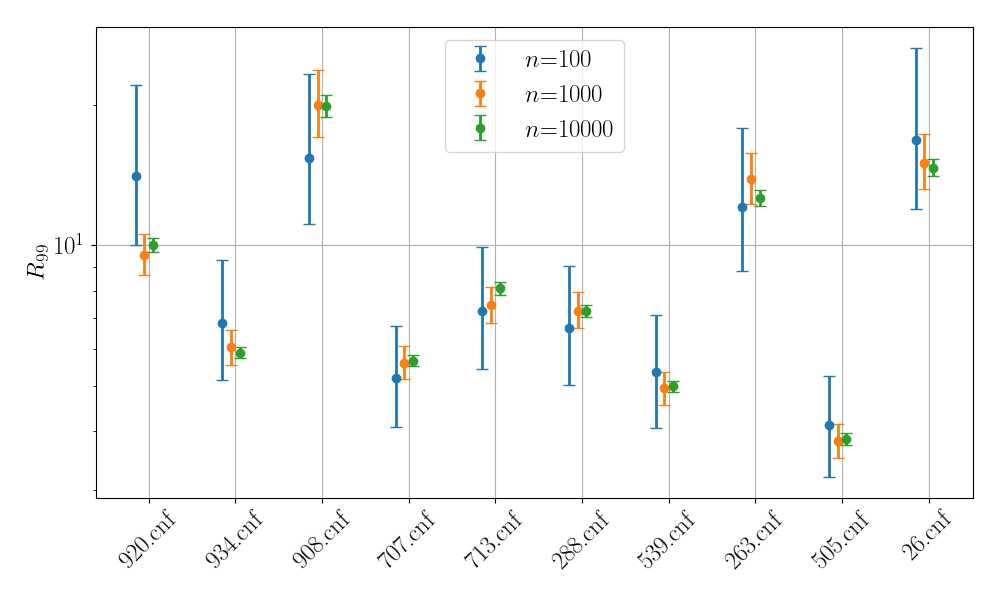}
        \caption{\texttt{CBS-k3-n100-m435-b10} }
        \label{fig: R99 CBS-k3-n100-m435-b10}
    \end{subfigure}

    \centering
    \begin{subfigure}{\linewidth}
    \includegraphics[width=\linewidth]{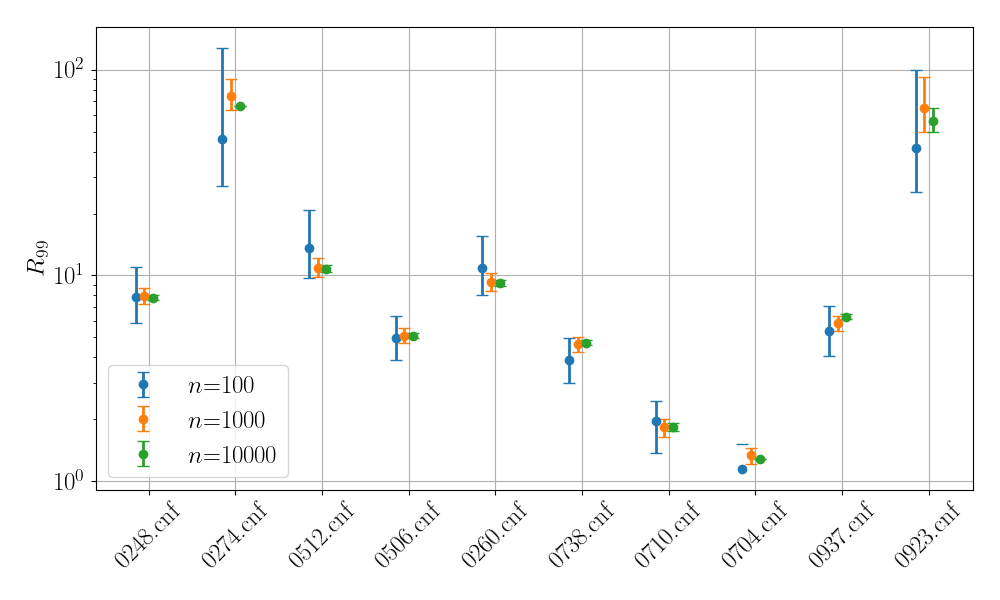}
    \caption{\texttt{uf100-430}}
    \label{fig: R99 uf100-430}
    \end{subfigure}

    \centering
    \begin{subfigure}{\linewidth}
        \includegraphics[width=\linewidth]{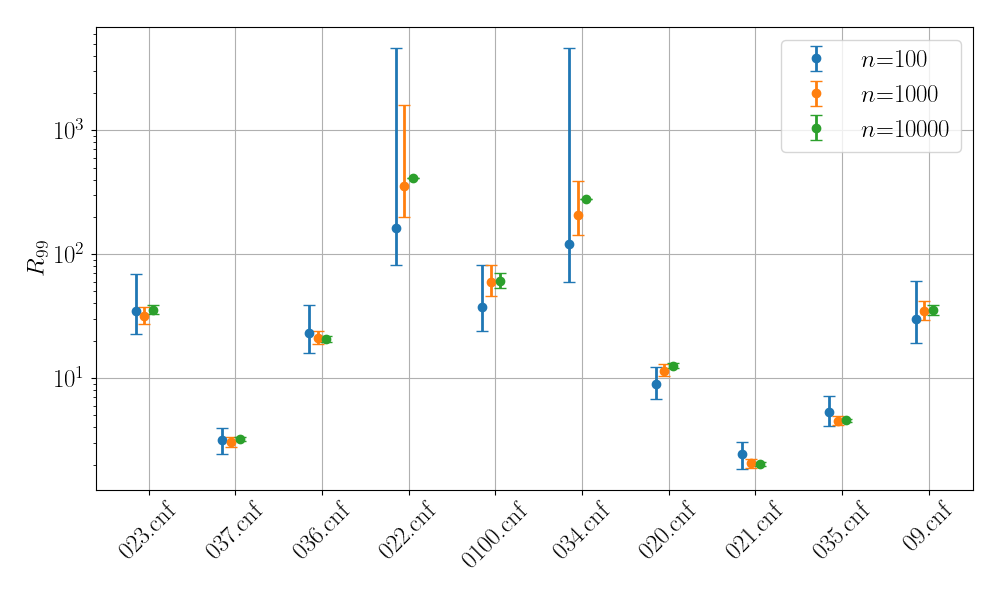}
        \caption{\texttt{uf250-1065}}
    \label{fig: R99 uf250-1065}
    \end{subfigure}

    \caption{Estimate of $R_{99}$ using $n \in \{ 100, 1000,$ 10,000$\}$ for three problem families. The horizontal axis shows the names of the problem instances within a family. The dots indicate the point estimates of $R_{99}$, and the lines and whiskers show the confidence intervals.}

    \label{fig: R99 all families}
\end{figure}

The results of the estimate of $R_{99}$ and its confidence interval across $n \in \{ 100, 1000,$ 10,000$\}$ are presented in Fig.~\ref{fig: R99 all families}. We make two important observations from this figure. First, for $n = 100$, the level of uncertainty in the estimate of $R_{99}$ can be very large, sometimes spanning even more than an order of magnitude, for example, 022.cnf and 034.cnf in the \texttt{uf250-1065} family. Increasing the number of repeats shrinks the confidence interval and improves the accuracy in the estimate, to the point that the uncertainty becomes negligible for $n =$ 10,000.

The second observation is that the point estimate of $R_{99}$ using a small number of repeats, for example, $n = 100$, could be incorrect. For the significant majority of  problems that were solved, the point estimates of $R_{99}$ with $n = 100$ repeats, indicated using blue dots, fall outside the confidence intervals of the estimates using \mbox{$n =$ 10,000} repeats, which contain the true value of $R_{99}$ with 99 percent confidence. Even in the case of $1000$ repeats, the point estimate of $R_{99}$ is sometimes incorrect. This suggests when only a small number of repeats is used, reporting a point estimate of $R_{99}$ can be misleading and the estimate should not be used in comparing different optimizers.

\begin{figure}
    \centering
    \includegraphics[width=\linewidth]{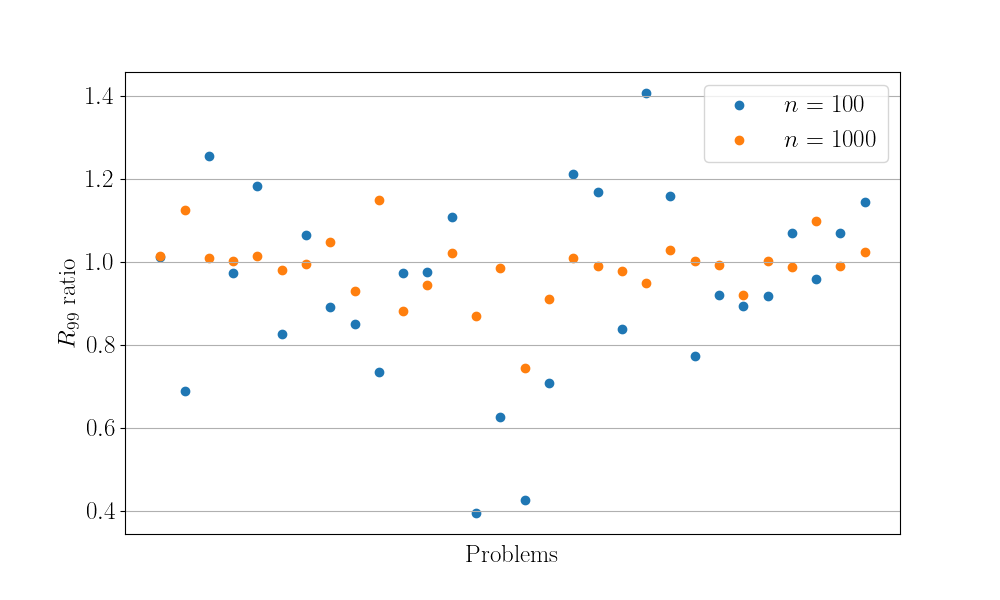}
    \caption{Ratio of the estimate of $R_{99}$ using 100 and 1000 repeats to the estimate using 10,000 repeats.}
    \label{fig: R99 ratio}
\end{figure}

To better illustrate how far the point estimates obtained using 100 and 1000 repeats can be from the true $R_{99}$ value, we plot the ratio of the estimate of $R_{99}$ when using 100 and 1000 repeats to its estimate when using 10,000 repeats (see Fig.~\ref{fig: R99 ratio}). Note that here, we use the estimate based on 10,000 repeats as a proxy for the true value of $R_{99}$ due to its high estimation accuracy. As shown in this figure, the point estimate using $n = 100$ repeats may underestimate $R_{99}$ by up to 60 percent or overestimate it by up to 40 percent for the studied problems. The estimation error is smaller when using 1000 repeats; the orange dots fall closer to the $ y= 1$ line compared to the blue dots.

\begin{figure}[h]
    \centering
    \vspace{0.5cm}
    \begin{subfigure}{\linewidth}
        \includegraphics[width=\linewidth]{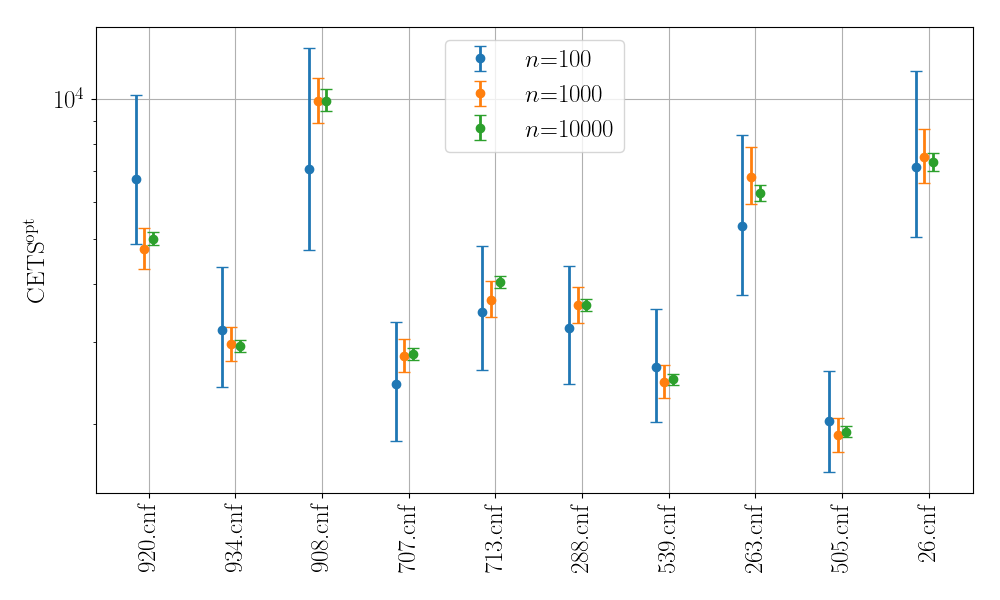}
        \caption{\texttt{CBS-k3-n100-m435-b10} }
        \label{fig: R99 CBS-k3-n100-m435-b10}
    \end{subfigure}

    \centering
    \begin{subfigure}{\linewidth}
    \includegraphics[width=\linewidth]{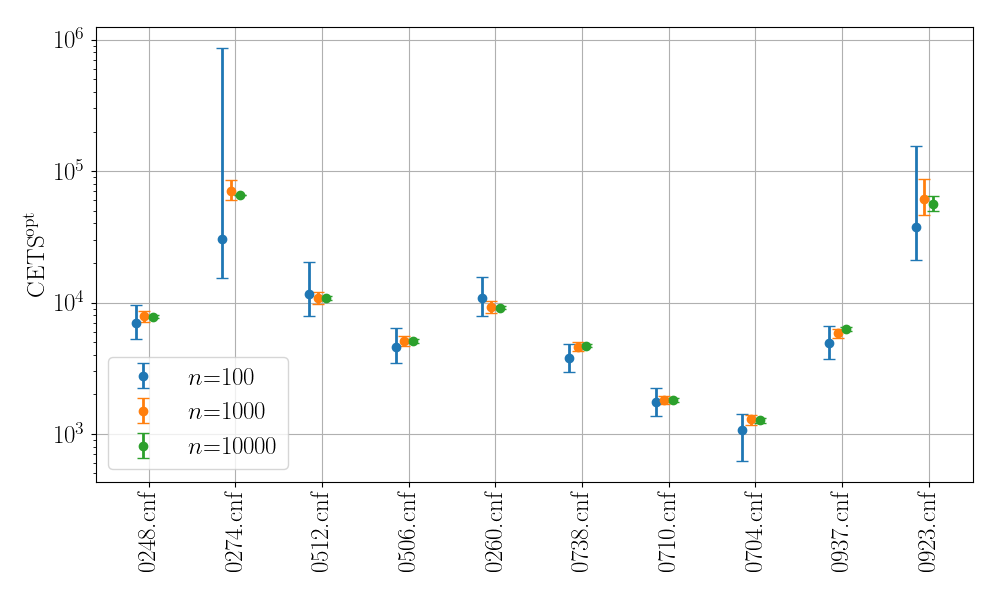}
    \caption{\texttt{uf100-430}}
    \label{fig: R99 uf100-430}
    \end{subfigure}

    \centering
    \begin{subfigure}{\linewidth}
        \includegraphics[width=\linewidth]{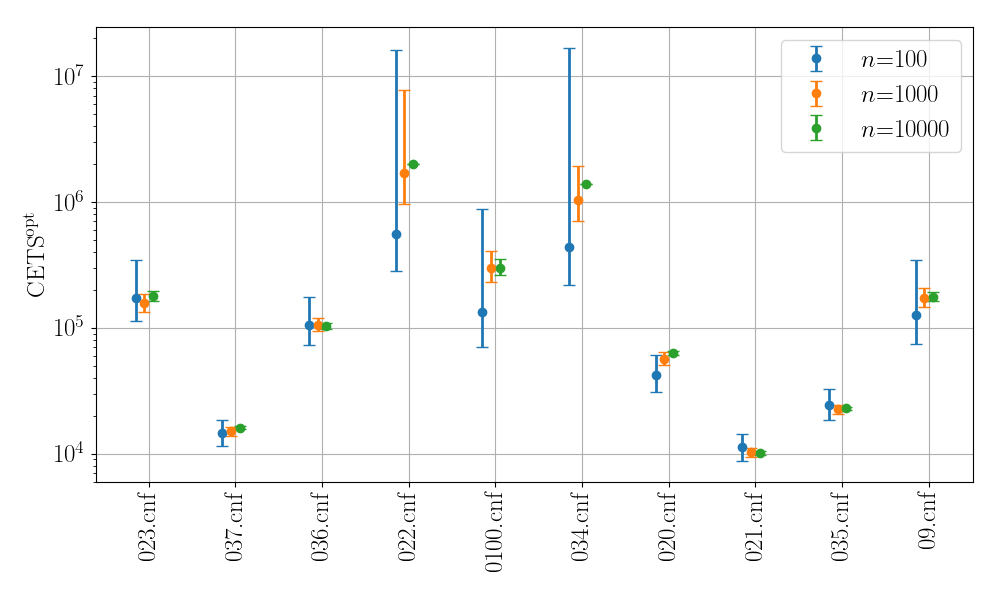}
        \caption{\texttt{uf250-1065}}
    \label{fig: R99 uf250-1065}
    \end{subfigure}

    \caption{Estimated values of $\CETSopt$ using $n \in \{ 100, 1000,$ 10,000$\}$ for three problem families. The horizontal axis shows the names of the problem instances within a family. The dots indicate the point estimates of the $\CETSopt$ values, and the lines and whiskers show the confidence intervals.}

    \label{fig: CETS all families}
\end{figure}

\begin{figure}[h]
    \centering
    \begin{subfigure}{\linewidth}
        \includegraphics[width=\linewidth]{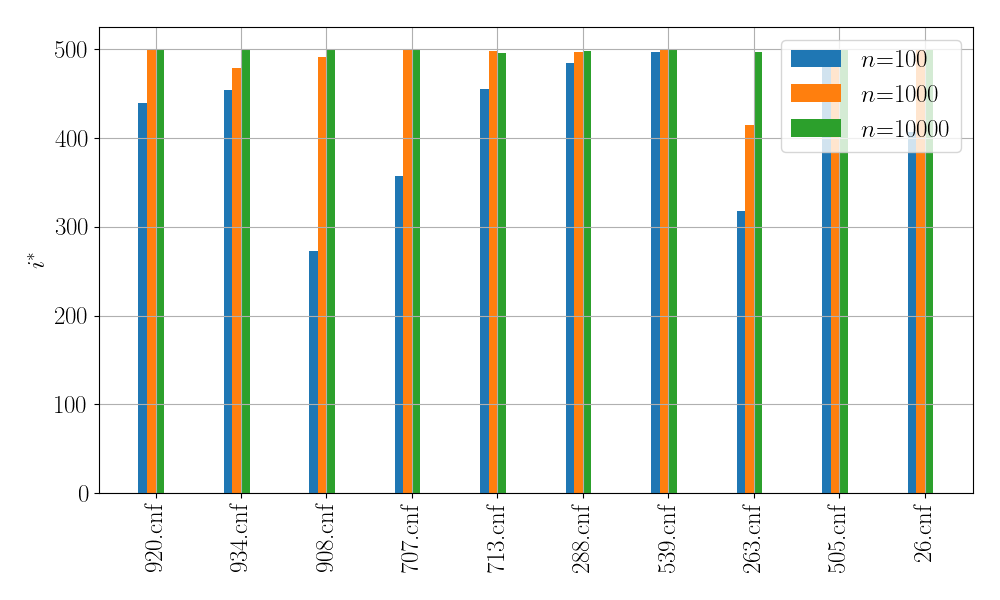}
        \caption{\texttt{CBS-k3-n100-m435-b10} }
        \label{fig: R99 CBS-k3-n100-m435-b10}
    \end{subfigure}

    \centering
    \begin{subfigure}{\linewidth}
    \includegraphics[width=\linewidth]{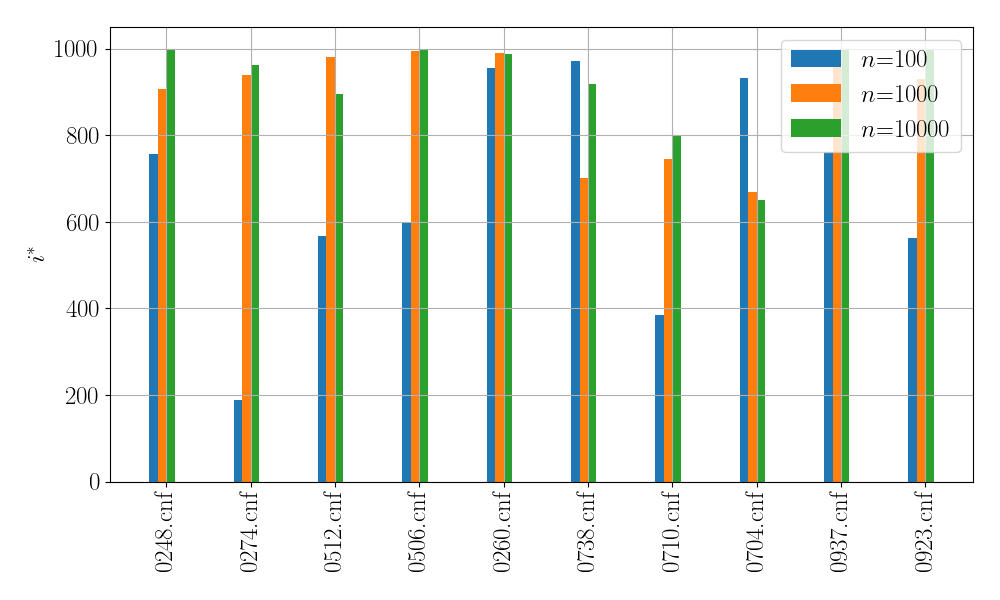}
    \caption{\texttt{uf100-430}}
    \label{fig: R99 uf100-430}
    \end{subfigure}

    \centering
    \begin{subfigure}{\linewidth}
        \includegraphics[width=\linewidth]{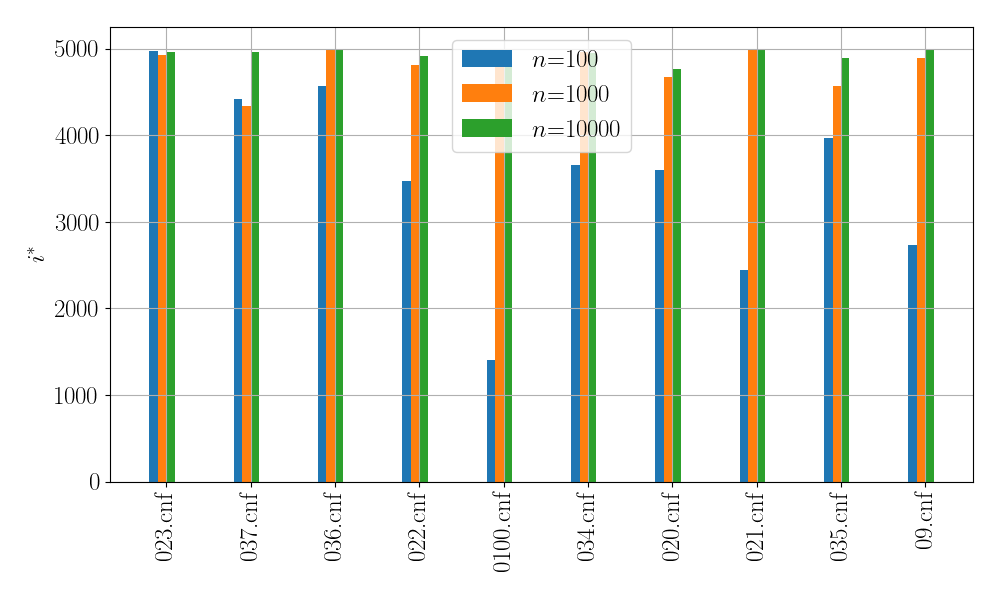}
        \caption{\texttt{uf250-1065}}
    \label{fig: R99 uf250-1065}
    \end{subfigure}

    \caption{The optimal number of iterations $i^*$ found using $n \in \{ 100, 1000,$ 10,000$\}$ for three problem families. The horizontal axis shows names of the problem instances within a family.}

    \label{fig: itr all families}
\end{figure}

We demonstrate the utility of our statistical analysis and the importance of choosing an appropriate number of repeats to use to determine $\CETSopt$. In the following experiment,
\begin{equation}
    \CETS = i R_{99},
\end{equation}
where $i$ is the number of WalkSAT iterations. This metric is often referred to as iterations-to-solution (ITS). To find the value of $\CETSopt$, we optimize over $i$ according to formulation~\eqref{eq: opt CEST}. The results for $\CETSopt$ and its confidence interval are shown in Fig.~\ref{fig: CETS all families}. The optimal number of iterations $i^*$ needed to find $\CETSopt$ is shown in Fig.~\ref{fig: itr all families}.

Similar to the results for $R_{99}$, the point estimate of $\CETSopt$ has significant uncertainty for $n = 100$ and may be a significant over- or underestimation of the true $\CETSopt$ value. It is evident from Fig.~\ref{fig: itr all families} that $i^*$  varies significantly as the number of repeats is increased. The value of $i^*$ determines the optimal strategy for running the optimizer to find the value of $\CETSopt$. A small value of $i^*$ indicates that the optimal strategy is to run WalkSAT in the fail-fast mode, that is, with a small number of iterations and restarting it frequently. On the other hand, larger values of $i^*$ indicate utilizing the optimizer in the patient mode, that is, executing
WalkSAT for a few long but highly successful runs. We can see that, for example, for problem 0274.cnf in the \texttt{uf100-430} family, the number of repeats affects the optimal strategy chosen for the solver. With $n = 100$ repeats, the optimal strategy is the fail-fast mode (i.e., restarting at about 200 iterations), whereas with $n  =$ 10,000 repeats, it is to run the solver in the patient mode.

\subsection{Number of Repeats and Hyperparameter Optimization}
The number of repeats also plays an important role in conducting a reliable hyperparameter optimization (HPO). If there is significant uncertainty in the metric's estimate, the HPO will not be able to converge to a truly optimal set of hyperparameters. This is because the HPO may identify a set of hyperparameters to be better than another set, while it is indeed worse.

\begin{table}[h]
\begin{subtable}{\linewidth}
    \centering
    \begin{tabular}{|c|c|c|c|c|}
        \hline
         $n$ & $p$ & $R_{99}$ & $\CIa$ of $R_{99}$ & $e(R_{99})$ \\ \hline
         100 & 0.09 $\pm$ 0.05 & 51.27 & (31.53, 125.94) & 1.46 \\
         1000 & 0.06 $\pm$ 0.02 & 68.88 & (52.05, 100.77) & 0.46 \\
         10,000 & 0.07 $\pm$ 0.01 & 62.22 & (54.13, 72.48) & 0.17 \\
         \hline
    \end{tabular}
    \caption{Effect of $n$ on the estimate of $p$ and $R_{99}$ for $w = 0.48$.}
    \label{tab: real data success w= 0.48}
\end{subtable}

\begin{subtable}{\linewidth}
    \centering
    \begin{tabular}{|c|c|c|c|c|}
        \hline
         $n$ & $p$ & $R_{99}$ & $\CIa$ of $R_{99}$ & $e(R_{99})$ \\ \hline
         100 & 0.10 $\pm$ 0.06 & 45.86 & (27.24, 127.27) & 1.78 \\
         1000 & 0.07 $\pm$ 0.02 & 66.75 & (50.80, 96.37) & 0.44 \\
         10,000 & 0.06 $\pm$ 0.00 & 69.44 & (69.44, 69.44) & 0.00 \\
         \hline
    \end{tabular}
    \caption{Effect of $n$ on the estimate of $p$ and $R_{99}$ for $w = 0.52$.}
    \label{tab: real data success w = 0.52}
\end{subtable}
\caption{Effect of $n$ on the results of hyperparameter optimization.}
\label{tab: real data success w}
\end{table}

To demonstrate this, we solve uf100-0658.cnf from the \texttt{uf100-430} family using two walk probability values $w = 0.48$ and $w = 0.52$. These values are a slight deviation from the default value of $w = 0.5$, and a hyperparameter optimizer may decide to explore lower or higher walk probabilities to improve the performance, given the results for $w = 0.48$ and $w = 0.52$. The experiment's results are presented in Table~\ref{tab: real data success w}.

Based on running HPO using only $n = 100$ repeats, the HPO algorithm chooses $w = 0.52$ over $w = 0.48$, as it provides about an $11$ percent improvement in $R_{99}$. However, using $n =$ 10,000 repeats, the opposite is the case: running the WalkSAT algorithm with $w = 0.48$ outperforms running it with $w = 0.52$ by about $11$ percent.

%% file: metric.bib
@article{nikhar2024all,
  title={All-to-all reconfigurability with sparse and higher-order Ising machines},
  author={Nikhar, Srijan and Kannan, Sidharth and Aadit, Navid Anjum and Chowdhury, Shuvro and Camsari, Kerem Y},
  journal={Nature Communications},
  volume={15},
  number={1},
  pages={8977},
  year={2024},
  publisher={Nature Publishing Group UK London}
}

@article{yamamoto2017coherent,
  title={Coherent Ising machines—optical neural networks operating at the quantum limit},
  author={Yamamoto, Yoshihisa and Aihara, Kazuyuki and Leleu, Timothee and Kawarabayashi, Ken-ichi and Kako, Satoshi and Fejer, Martin and Inoue, Kyo and Takesue, Hiroki},
  journal={npj Quantum Information},
  volume={3},
  number={1},
  pages={49},
  year={2017},
  publisher={Nature Publishing Group UK London}
}

@book{biere2009handbook,
  title={Handbook of satisfiability},
  author={Biere, Armin and Heule, Marijn and van Maaren, Hans},
  volume={185},
  year={2009},
  publisher={IOS press}
}

@article{hoos2000local,
  title={Local search algorithms for SAT: An empirical evaluation},
  author={Hoos, Holger H and St{\"u}tzle, Thomas},
  journal={Journal of Automated Reasoning},
  volume={24},
  number={4},
  pages={421--481},
  year={2000},
  publisher={Springer}
}

@misc{WalkSAT,
  title = {WalkSAT Solver},
  howpublished = {\url{https://gitlab.com/HenryKautz/Walksat/-/archive/master/Walksat-master.tar.gz}},
}

@misc{SATLIB,
  title = {SATLIB - Benchmark Problems},
  howpublished = {\url{https://www.cs.ubc.ca/~hoos/SATLIB/benchm.html}},
}

@article{neira2025benchmarking,
  title={Benchmarking the operation of quantum heuristics and ising machines: scoring parameter setting strategies on optimization applications},
  author={Neira, David E Bernal and Brown, Robin and Sathe, Pratik and Wudarski, Filip and Pavone, Marco and Rieffel, Eleanor and Venturelli, Davide},
  journal={Quantum Machine Intelligence},
  volume={7},
  number={2},
  pages={86},
  year={2025}
}

@article{beiranvand2017best,
  title={Best practices for comparing optimization algorithms},
  author={Beiranvand, Vahid and Hare, Warren and Lucet, Yves},
  journal={Optimization and Engineering},
  volume={18},
  number={4},
  pages={815--848},
  year={2017},
  publisher={Springer}
}

@article{goto2021high,
  title={High-performance combinatorial optimization based on classical mechanics},
  author={Goto, Hayato and Endo, Kotaro and Suzuki, Masaru and Sakai, Yoshisato and Kanao, Taro and Hamakawa, Yohei and Hidaka, Ryo and Yamasaki, Masaya and Tatsumura, Kosuke},
  journal={Science Advances},
  volume={7},
  number={6},
  pages={eabe7953},
  year={2021},
  publisher={American Association for the Advancement of Science}
}

@article{patel2022logically,
  title={Logically synthesized and hardware-accelerated restricted {B}oltzmann machines for combinatorial optimization and integer factorization},
  author={Patel, Saavan and Canoza, Philip and Salahuddin, Sayeef},
  journal={Nature Electronics},
  volume={5},
  number={2},
  pages={92--101},
  year={2022},
  publisher={Nature Publishing Group}
}

@article{si2024energy,
  title={Energy-efficient superparamagnetic {I}sing machine and its application to traveling salesman problems},
  author={Si, Jia and Yang, Shuhan and Cen, Yunuo and Chen, Jiaer and Huang, Yingna and Yao, Zhaoyang and Kim, Dong-Jun and Cai, Kaiming and Yoo, Jerald and Fong, Xuanyao and others},
  journal={Nature Communications},
  volume={15},
  number={1},
  pages={3457},
  year={2024},
  publisher={Nature Publishing Group UK London}
}

@article{cilasun20243sat,
  title={{3SAT} on an all-to-all-connected {CMOS} {I}sing solver chip},
  author={C{\i}lasun, H{\"u}srev and Zeng, Ziqing and Kumar, Abhimanyu and Lo, Hao and Cho, William and Moy, William and Kim, Chris H and Karpuzcu, Ulya R and Sapatnekar, Sachin S},
  journal={Scientific reports},
  volume={14},
  number={1},
  pages={10757},
  year={2024},
  publisher={Nature Publishing Group UK London}
}

@inproceedings{zhang2022qubrim,
  title={Qubrim: A {CMOS} compatible resistively-coupled {I}sing machine with quantized nodal interactions},
  author={Zhang, Yiqiao and Vengalam, Uday Kumar Reddy and Sharma, Anshujit and Huang, Michael and Ignjatovic, Zeljko},
  booktitle={Proceedings of the 41st IEEE/ACM International Conference on Computer-Aided Design},
  pages={1--8},
  year={2022}
}

@article{bhattacharya2024computing,
  title={Computing high-degree polynomial gradients in memory},
  author={Bhattacharya, Tinish and Hutchinson, George H and Pedretti, Giacomo and Sheng, Xia and Ignowski, Jim and Van Vaerenbergh, Thomas and Beausoleil, Ray and Strachan, John Paul and Strukov, Dmitri B},
  journal={Nature Communications},
  volume={15},
  number={1},
  pages={8211},
  year={2024},
  publisher={Nature Publishing Group UK London}
}

@article{moy20221,
  title={A 1,968-node coupled ring oscillator circuit for combinatorial optimization problem solving},
  author={Moy, William and Ahmed, Ibrahim and Chiu, Po-wei and Moy, John and Sapatnekar, Sachin S and Kim, Chris H},
  journal={Nature Electronics},
  volume={5},
  number={5},
  pages={310--317},
  year={2022},
  publisher={Nature Publishing Group UK London}
}

@inproceedings{hizzani2024memristor,
  title={Memristor-based hardware and algorithms for higher-order {H}opfield optimization solver outperforming quadratic {I}sing machines},
  author={Hizzani, Mohammad and Heittmann, Arne and Hutchinson, George and Dobrynin, Dmitrii and Van Vaerenbergh, Thomas and Bhattacharya, Tinish and Renaudineau, Adrien and Strukov, Dmitri and Strachan, John Paul},
  booktitle={2024 IEEE International Symposium on Circuits and Systems (ISCAS)},
  pages={1--5},
  year={2024},
  organization={IEEE}
}

@article{pedretti2025solving,
  title={Solving {B}oolean satisfiability problems with resistive content addressable memories},
  author={Pedretti, Giacomo and B{\"o}hm, Fabian and Bhattacharya, Tinish and Heittman, Arne and Zhang, Xiangyi and Hizzani, Mohammad and Hutchinson, George and Kwon, Dongseok and Moon, John and Valiante, Elisabetta and others},
  journal={arXiv preprint arXiv:2501.07733},
  year={2025}
}

@article{bianchi2009survey,
  title={A survey on metaheuristics for stochastic combinatorial optimization},
  author={Bianchi, Leonora and Dorigo, Marco and Gambardella, Luca Maria and Gutjahr, Walter J},
  journal={Natural Computing},
  volume={8},
  pages={239--287},
  year={2009},
  publisher={Springer}
}

@article{schuman2022opportunities,
  title={Opportunities for neuromorphic computing algorithms and applications},
  author={Schuman, Catherine D and Kulkarni, Shruti R and Parsa, Maryam and Mitchell, J Parker and Date, Prasanna and Kay, Bill},
  journal={Nature Computational Science},
  volume={2},
  number={1},
  pages={10--19},
  year={2022},
  publisher={Nature Publishing Group US New York}
}

@article{verma2019memory,
  title={In-memory computing: Advances and prospects},
  author={Verma, Naveen and Jia, Hongyang and Valavi, Hossein and Tang, Yinqi and Ozatay, Murat and Chen, Lung-Yen and Zhang, Bonan and Deaville, Peter},
  journal={IEEE Solid-State Circuits Magazine},
  volume={11},
  number={3},
  pages={43--55},
  year={2019},
  publisher={IEEE}
}

@book{horowitz2019quantum,
  title={Quantum computing: progress and prospects},
  author={Horowitz, Mark and Grumbling, Emily},
  year={2019},
  publisher={National Academies Press},
  journal={},
  volume={},
  number={},
  publisher = {National Academies Press}
}

@article{goto2019combinatorial,
  title={Combinatorial optimization by simulating adiabatic bifurcations in nonlinear {H}amiltonian systems},
  author={Goto, Hayato and Tatsumura, Kosuke and Dixon, Alexander R},
  journal={Science advances},
  volume={5},
  number={4},
  pages={eaav2372},
  year={2019},
  publisher={American Association for the Advancement of Science}
}

@article{tsukamoto2017accelerator,
  title={An accelerator architecture for combinatorial optimization problems},
  author={Tsukamoto, Sanroku and Takatsu, Motomu and Matsubara, Satoshi and Tamura, Hirotaka},
  journal={Fujitsu Sci. Tech. J},
  volume={53},
  number={5},
  pages={8--13},
  year={2017}
}

@article{morita2008mathematical,
  title={Mathematical foundation of quantum annealing},
  author={Morita, Satoshi and Nishimori, Hidetoshi},
  journal={Journal of Mathematical Physics},
  volume={49},
  number={12},
  year={2008},
  publisher={AIP Publishing}
}

@article{barrero2015study,
  title={A study on {K}oza’s performance measures},
  author={Barrero, David F and Castano, Bonifacio and R-Moreno, Mar{\'\i}a D and Camacho, David},
  journal={Genetic Programming and Evolvable Machines},
  volume={16},
  pages={327--349},
  year={2015},
  publisher={Springer}
}

@inproceedings{keijzer2001adaptive,
  title={Adaptive logic programming},
  author={Keijzer, M and Babovic, V and Ryan, C and O'Neill, M and Cattolico, M},
  booktitle={Proceedings of the 3rd Annual Conference on Genetic and Evolutionary Computation},
  pages={42--49},
  year={2001}
}

@inproceedings{christensen2002analysis,
  title={An analysis of {K}oza’s computational effort statistic for genetic programming},
  author={Christensen, Steffen and Oppacher, Franz},
  booktitle={Proceedings of 5th European Conference on Genetic Programming (EuroGP)},
  pages={182--191},
  year={2002},
  organization={Springer}
}

@inproceedings{angeline1996investigation,
  title={An investigation into the sensitivity of genetic programming to the frequency of leaf selection during subtree crossover},
  author={Angeline, Peter J},
  booktitle={Proceedings of the 1st annual conference on genetic programming},
  pages={21--29},
  year={1996}
}

@article{perdomo2019readiness,
  title={Readiness of quantum optimization machines for industrial applications},
  author={Perdomo-Ortiz, Alejandro and Feldman, Alexander and Ozaeta, Asier and Isakov, Sergei V and Zhu, Zheng and O’Gorman, Bryan and Katzgraber, Helmut G and Diedrich, Alexander and Neven, Hartmut and de Kleer, Johan and others},
  journal={Physical Review Applied},
  volume={12},
  number={1},
  pages={014004},
  year={2019},
  publisher={APS}
}

@article{albash2018demonstration,
  title={Demonstration of a scaling advantage for a quantum annealer over simulated annealing},
  author={Albash, Tameem and Lidar, Daniel A},
  journal={Physical Review X},
  volume={8},
  number={3},
  pages={031016},
  year={2018},
  publisher={APS}
}

@inproceedings{cai2013improving,
  title={Improving {WalkSAT} for random k-satisfiability problem with $k > 3$},
  author={Cai, Shaowei and Su, Kaile and Luo, Chuan},
  booktitle={Proceedings of the AAAI Conference on Artificial Intelligence},
  volume={27},
  number={1},
  pages={145--151},
  year={2013}
}

@inproceedings{beeson2015trac,
  title={TRAC-IK: An open-source library for improved solving of generic inverse kinematics},
  author={Beeson, Patrick and Ames, Barrett},
  booktitle={2015 IEEE-RAS 15th International Conference on Humanoid Robots (Humanoids)},
  pages={928--935},
  year={2015},
  organization={IEEE}
}

@article{chmiela2021learning,
  title={Learning to schedule heuristics in branch and bound},
  author={Chmiela, Antonia and Khalil, Elias and Gleixner, Ambros and Lodi, Andrea and Pokutta, Sebastian},
  journal={Advances in Neural Information Processing Systems},
  volume={34},
  pages={24235--24246},
  year={2021}
}

@article{mcgeoch2024not,
  title={How NOT to Fool the Masses When Giving Performance Results for Quantum Computers},
  author={McGeoch, Catherine},
  journal={arXiv preprint arXiv:2411.08860},
  year={2024}
}

@article{kowalsky20223,
  title={3-regular three-{XORSAT} planted solutions benchmark of classical and quantum heuristic optimizers},
  author={Kowalsky, Matthew and Albash, Tameem and Hen, Itay and Lidar, Daniel A},
  journal={Quantum Science and Technology},
  volume={7},
  number={2},
  pages={025008},
  year={2022},
  publisher={IOP Publishing}
}

@article{ronnow2014defining,
  title={Defining and detecting quantum speedup},
  author={R{\o}nnow, Troels F and Wang, Zhihui and Job, Joshua and Boixo, Sergio and Isakov, Sergei V and Wecker, David and Martinis, John M and Lidar, Daniel A and Troyer, Matthias},
  journal={science},
  volume={345},
  number={6195},
  pages={420--424},
  year={2014},
  publisher={American Association for the Advancement of Science}
}

@book{koza1994genetic,
  title={Genetic programming II: automatic discovery of reusable programs},
  author={Koza, John R},
  year={1994},
  publisher={MIT Press}
}

@incollection{collet2008stochastic,
  title={Stochastic optimization algorithms},
  author={Collet, Pierre and Rennard, Jean-Philippe},
  booktitle={Intelligent information technologies: Concepts, methodologies, tools, and applications},
  pages={1121--1137},
  year={2008},
  publisher={IGI Global}
}

@article{gonccalves2012sample,
  title={Sample size for estimating a binomial proportion: comparison of different methods},
  author={Gon{\c{c}}alves, Luzia and de Oliveira, M Ros{\'a}rio and Pascoal, Cl{\'a}udia and Pires, Ana},
  journal={Journal of Applied Statistics},
  volume={39},
  number={11},
  pages={2453--2473},
  year={2012},
  publisher={Taylor \& Francis}
}

@article{krishnamoorthy2007some,
  title={Some properties of the exact and score methods for binomial proportion and sample size calculation},
  author={Krishnamoorthy, K and Peng, Jie},
  journal={Communications in Statistics—Simulation and Computation},
  volume={36},
  number={6},
  pages={1171--1186},
  year={2007},
  publisher={Taylor \& Francis}
}

@article{brown2001interval,
  title={Interval estimation for a binomial proportion},
  author={Brown, Lawrence D and Cai, T Tony and DasGupta, Anirban},
  journal={Statistical science},
  volume={16},
  number={2},
  pages={101--133},
  year={2001},
  publisher={Institute of Mathematical Statistics}
}

@article{brown2002confidence,
  title={Confidence intervals for a binomial proportion and asymptotic expansions},
  author={Brown, Lawrence D and Cai, T Tony and DasGupta, Anirban},
  journal={The Annals of Statistics},
  volume={30},
  number={1},
  pages={160--201},
  year={2002},
  publisher={Institute of Mathematical Statistics}
}

@article{anderson23,
author = {Per Gösta Andersson},
title = {The {W}ald Confidence Interval for a Binomial $p$ as an Illuminating “Bad” Example},
journal = {The American Statistician},
volume = {77},
number = {4},
pages = {443--448},
year = {2023},
publisher = {ASA Website},
doi = {10.1080/00031305.2023.2183257},
}
